\documentclass{article}

  \usepackage[main, final]{neurips_2026}

\usepackage[utf8]{inputenc} 
\usepackage[T1]{fontenc}    
\usepackage{hyperref}       
\usepackage{url}            
\usepackage{booktabs}       
\usepackage{amsfonts}       
\usepackage{nicefrac}       
\usepackage{microtype}      
\usepackage{xcolor}         
\usepackage[table]{xcolor}

\usepackage{xspace}

\usepackage{amsmath,amssymb}
\usepackage{graphicx}
\usepackage{multirow}
\usepackage{subcaption}
\usepackage{siunitx}
\usepackage{float}
\usepackage{tikz}
\usetikzlibrary{positioning,arrows.meta,calc,fit,shapes.geometric,decorations.pathmorphing,decorations.markings}

\newcommand{\hmcBalAcc}{0.764 $\pm$ 0.002}
\newcommand{\hmcFtxt}{0.775 $\pm$ 0.004}
\newcommand{\hmcKappa}{0.711 $\pm$ 0.004}
\newcommand{\isrucBalAcc}{0.806 $\pm$ 0.001}
\newcommand{\isrucFtxt}{0.829 $\pm$ 0.004}
\newcommand{\isrucKappa}{0.780 $\pm$ 0.005}
\newcommand{\seedvBalAcc}{0.492 $\pm$ 0.009}
\newcommand{\seedvFtxt}{0.468 $\pm$ 0.019}
\newcommand{\seedvKappa}{0.355 $\pm$ 0.016}
\newcommand{\sienaBalAcc}{0.776 $\pm$ 0.010}
\newcommand{\sienaAUROC}{0.883 $\pm$ 0.012}
\newcommand{\sienaAUCPR}{0.383 $\pm$ 0.036}
\newcommand{\tuabBalAcc}{0.819 $\pm$ 0.004}
\newcommand{\tuabAUROC}{0.892 $\pm$ 0.008}
\newcommand{\tuabAUCPR}{0.896 $\pm$ 0.006}
\newcommand{\tuevBalAcc}{0.715 $\pm$ 0.013}
\newcommand{\tuevFtxt}{0.806 $\pm$ 0.010}
\newcommand{\tuevKappa}{0.625 $\pm$ 0.019}

\newcommand{\OneDhmcBalAcc}{0.745 $\pm$ 0.005}
\newcommand{\OneDisrucBalAcc}{0.801 $\pm$ 0.001}
\newcommand{\OneDmumtazBalAcc}{0.895 $\pm$ 0.006}
\newcommand{\OneDsienaBalAcc}{0.745 $\pm$ 0.016}
\newcommand{\OneDtuabBalAcc}{0.826 $\pm$ 0.004}
\newcommand{\OneDtuevBalAcc}{0.559 $\pm$ 0.021}
\newcommand{\OneDIdentityhmcBalAcc}{0.727 $\pm$ 0.007}
\newcommand{\OneDIdentityisrucBalAcc}{0.789 $\pm$ 0.004}
\newcommand{\OneDIdentitymumtazBalAcc}{0.896 $\pm$ 0.003}
\newcommand{\OneDIdentitysienaBalAcc}{0.710 $\pm$ 0.038}
\newcommand{\OneDIdentitytuabBalAcc}{0.800 $\pm$ 0.004}
\newcommand{\OneDIdentitytuevBalAcc}{0.675 $\pm$ 0.023}
\newcommand{\TwoDhmcBalAcc}{0.749 $\pm$ 0.005}
\newcommand{\TwoDisrucBalAcc}{0.800 $\pm$ 0.006}
\newcommand{\TwoDmumtazBalAcc}{0.909 $\pm$ 0.004}
\newcommand{\TwoDsienaBalAcc}{0.723 $\pm$ 0.022}
\newcommand{\TwoDtuabBalAcc}{0.817 $\pm$ 0.004}
\newcommand{\TwoDtuevBalAcc}{0.519 $\pm$ 0.016}

\newcommand{\OneDbcicivaBalAccSweep}{0.420 $\pm$ 0.013}
\newcommand{\OneDmatBalAccSweep}{0.708 $\pm$ 0.036}
\newcommand{\OneDseedvBalAccSweep}{0.438 $\pm$ 0.013}
\newcommand{\OneDIdentitybcicivaBalAccSweep}{0.458 $\pm$ 0.029}
\newcommand{\OneDIdentitymatBalAccSweep}{0.528 $\pm$ 0.021}
\newcommand{\OneDIdentityseedvBalAccSweep}{0.485 $\pm$ 0.009}
\newcommand{\bcicivaBalAccSweep}{0.570 $\pm$ 0.008}
\newcommand{\bcicivaFtxtSweep}{0.567 $\pm$ 0.008}
\newcommand{\bcicivaKappaSweep}{0.427 $\pm$ 0.011}
\newcommand{\matBalAccSweep}{0.840 $\pm$ 0.022}
\newcommand{\matAUROCSweep}{0.912 $\pm$ 0.022}
\newcommand{\matAUCPRSweep}{0.806 $\pm$ 0.038}
\newcommand{\mumtazBalAccSweep}{1.000 $\pm$ 0.000}
\newcommand{\mumtazAUROCSweep}{1.000 $\pm$ 0.000}
\newcommand{\mumtazAUCPRSweep}{1.000 $\pm$ 0.000}
\newcommand{\TwoDbcicivaBalAccSweep}{0.468 $\pm$ 0.007}
\newcommand{\TwoDmatBalAccSweep}{0.715 $\pm$ 0.067}
\newcommand{\TwoDseedvBalAccSweep}{0.470 $\pm$ 0.011}

\newcommand{\vect}[1]{\boldsymbol{#1}}
\newcommand{\R}{\mathbb{R}}

\newcommand{\ie}{\textit{i.e.}}

\newcommand{\model}{B[FM]$^2$\xspace}   

\title{\model: Brain Foundation Model via\\Flow Matching with SplitUNet}

\author{%
  Jaedong Hwang$^1$
  \quad
  Kathleen Zhang$^1$
  \quad
  Wei Dai$^1$
  \quad
  Konstantinos Kontras$^{1,2}$
  \\
  \textbf{Maarten Vanmarcke$^2$}
   \quad 
  \textbf{Maarten De Vos$^2$}
  \quad
    \textbf{Ila Fiete$^1$}
    \quad
  \textbf{Paul Pu Liang$^1$}
  \\
  $^1$Massachusetts Institute of Technology \quad
  $^2$KU Leuven
}

\begin{document}

\maketitle

\begin{abstract}

EEG foundation models can learn generalizable representations from large-scale EEG corpora to enable single-backbone transfer across diverse clinical and brain-computer interface tasks.
Existing models typically discretize the continuous multi-channel EEG waveform into patches or codebook tokens and train a transformer with masked self-supervision. 
Recognizing that this discretization fragments continuous brain rhythms and obscures fine-grained temporal dynamics, we present \model (Brain Foundation Model via Flow Matching), whose inductive bias aligns with the data by pretraining directly on the raw signal using continuous-time flow matching without patches, tokenization, or masking.
However, multi-channel EEG signals pose an architectural challenge for flow matching:
time is densely sampled and highly autocorrelated (thousands of timepoints), while the electrode axis is short (tens of channels) at distinct scalp positions.
To address this time-electrode asymmetry, we introduce SplitUNet, a velocity network that factorizes each block into separate 1D temporal and 1D electrode convolutions and downsamples only along time, preserving electrode topology throughout the hierarchy.
\model sets a new state of the art on $7$ of $9$ standard downstream EEG classification tasks, using a pretraining budget of only $36{,}895$ segments ($\approx 307$\,h), 1--2 orders of magnitude ($\approx 30\times$) less than required by existing EEG foundation models. Further, it generates synthetic EEGs that two board-certified neurologists cannot distinguish from brain data (Cohen's $\kappa = -0.096$)\footnote{\url{https://jd730.github.io/projects/BFM2}}.
\end{abstract}


\section{Introduction}
\label{sec:intro}

Brain foundation models (BFMs) can enable general-purpose EEG (electroencephalogram) analysis without task-specific supervision, successfully transferring a single pretrained backbone across diverse tasks like seizure detection and sleep staging. Driven by large open corpora~\citep{obeid2016temple} and the adoption of vision and language architectures~\citep{he2022masked,liu2023visual,radford2019language}, the field has seen rapid progress.
However, to leverage these architectures, existing BFMs~\citep{jiang2024large,kostas2021bendr,ouahidi2025reve,yang2023biot} have universally adopted a rigid training paradigm imported directly from computer vision: continuous neural waveforms are forcibly discretized into patches or learned codebooks, and a Transformer~\citep{vaswani2017attention} is trained via discrete masked reconstruction or contrastive prediction.

However, for scalp EEG, this discretization introduces a fundamental representational mismatch.
Brain rhythms are continuous, dynamic processes, yet existing approaches fragment them into arbitrary tokens with artificial boundaries.
Consequently, the self-supervised objective is forced to model transitions between discrete chunks rather than capturing the underlying continuous temporal evolution of the neural signal~\citep{rubanova2019latent,zeng2023transformers}.
Furthermore, random patch masking obscures the fine-grained temporal variations that are crucial for downstream clinical tasks~\citep{dong2023simmtm}.

To resolve this mismatch, we introduce \model (Brain Foundation Model via Flow Matching), which eliminates the discretization entirely.
We directly pretrain on the continuous multi-channel waveform via flow matching~\citep{lipman2023flow,tong2024improving}.
By corrupting the signal across all electrodes and timepoints simultaneously along a trajectory from Gaussian noise to data, the model learns a denoising vector field that inherently captures the continuous multi-scale dynamics of the brain.
While continuous-time generative frameworks have yielded strong semantic representations in vision~\citep{baranchuk2022labelefficient,tang2023emergent,xu2023open}, porting them to multi-channel EEG presents a distinct architectural challenge: 
dense temporal dynamics are observed only through the small, anatomically constrained set of scalp electrodes that the EEG montage permits.

\begin{figure}[t!]
  \centering
  \includegraphics[width=0.95\linewidth]{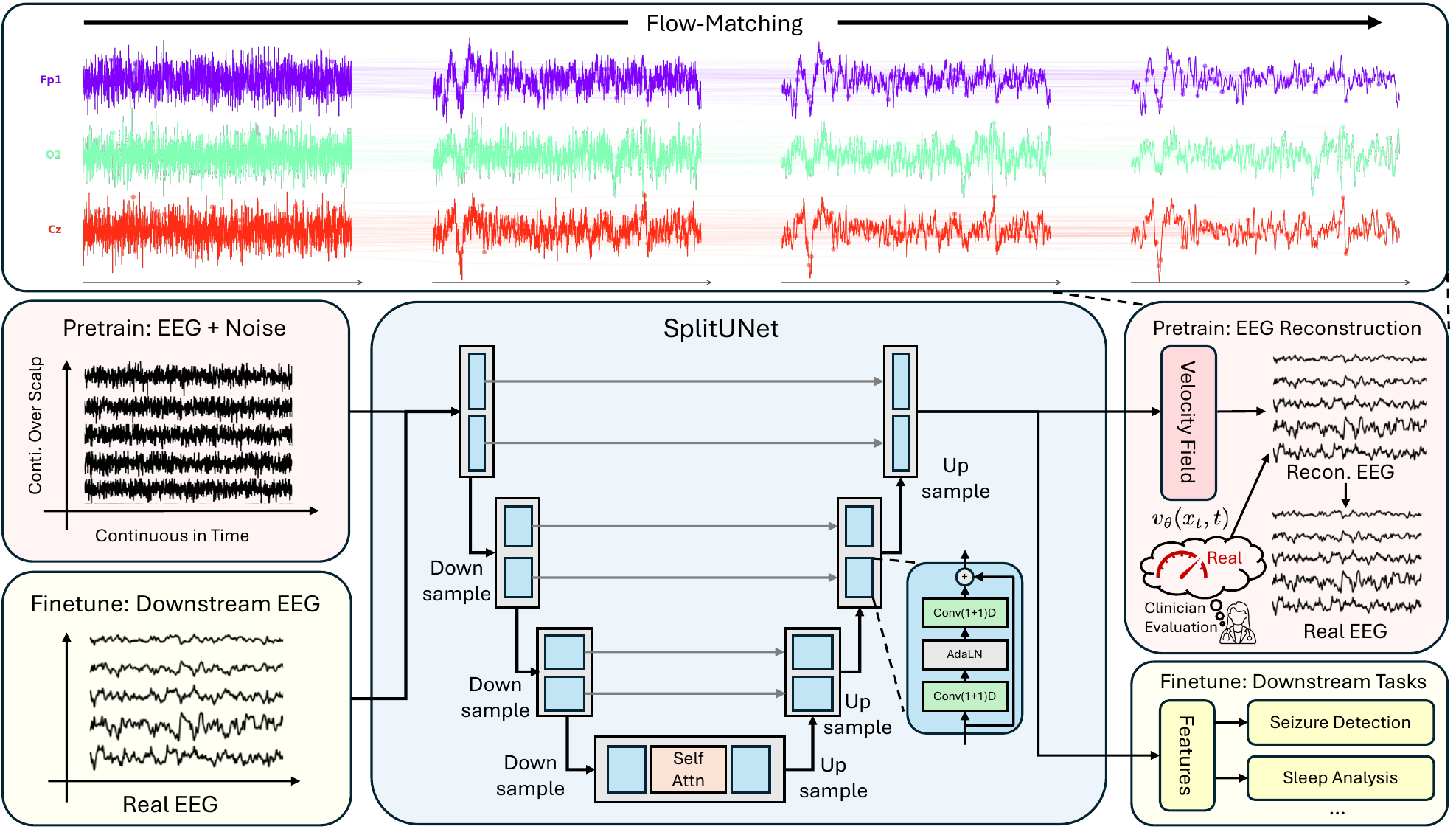}
  \caption{
  \textbf{Continuous-time generative pretraining for EEG.}
\textbf{(Top)}
The flow matching process maps Gaussian noise ($t=0.0$) to continuous, multi-channel EEG signals ($t=1.0$) along a continuous trajectory.
\textbf{(Left/Center)}
SplitUNet, the velocity network, reads the interpolant $\vect{x}_t$ together with $t$ and predicts the velocity $v_\theta(\vect{x}_t, t)$ that points toward the data endpoint.
Every spatiotemporal convolution inside it is factorized into a 1D convolution along time followed by a 1D convolution along electrodes, and down/upsampling acts on the time axis only, so the $E$ electrodes survive at full resolution from input to output.
\textbf{(Right)} Integrating the velocity field from noise generates new EEG, which two board-certified neurologists could not reliably separate from real recordings.
}
  \label{fig:overview}
\vspace{-0.35cm}
\end{figure}

To address this challenge, we propose \emph{SplitUNet}, a velocity-network backbone designed for multi-channel EEG, where dense temporal dynamics are observed through a small, fixed set of electrodes.
Each convolutional block factorizes spatiotemporal mixing into a 1D temporal convolution followed by a 1D electrode convolution;
the encoder downsamples only along time, preserving the electrode dimension at every layer; and the network operates directly on the raw waveform---no patches, codebooks, or masking.
With a continuous-time flow-matching objective, this backbone generates EEG by integrating a velocity field $v_\theta(\vect{x}_t, t)$ from Gaussian noise to the data manifold, producing a single velocity prediction over the full multi-channel signal at each step.
Figure~\ref{fig:overview} depicts the pretraining-and-finetuning pipeline.

\model achieves state-of-the-art performance on $7$ of $9$ downstream tasks, outperforming prior models by up to $7.4$\,pp.
Crucially, this is achieved with a pretraining set of only $36{,}895$ segments ($\approx 307$\,h), one to two orders of magnitude smaller than corpora used by recent EEG foundation models.
This establishes continuous-time flow matching as a markedly more sample-efficient pretraining framework than masked-token reconstruction.
Furthermore, the resulting generative model produces highly realistic EEG that board-certified neurologists failed to distinguish from real TUEG data.

To summarize, the contributions of this paper are four-fold:
\begin{itemize}
\setlength{\itemsep}{1pt}\setlength{\parsep}{0pt}\setlength{\topsep}{2pt}\setlength{\partopsep}{0pt}
\item We introduce a continuous-time flow matching paradigm~\citep{lipman2023flow,tong2024improving} for EEG foundation models, replacing the standard discrete masked-token framework. By operating directly on the raw multi-channel waveform, this generative pretext proves highly sample-efficient, requiring pretraining on only $36{,}895$ segments.
\item We propose \emph{SplitUNet}, an architecture tailored for continuous brain signals that factorizes temporal and electrode convolutions, preserving electrode topology end-to-end without relying on patches.
\item We demonstrate that \model achieves state-of-the-art performance on $7$ of $9$ tasks across a diverse downstream suite.
\item Simultaneously, \model can generate synthetic EEGs, which are demonstrated to be clinically indistinguishable from real TUEG data to board-certified neurologists.
\end{itemize}
\section{\model: Architecture and Pretraining}
\label{sec:method}

\subsection{Preliminary: Flow Matching}\label{subsec:preliminaries}

We pretrain a velocity network $v_\theta$ on multi-channel EEG waveforms $\vect{x}\in\R^{E\times T}$, where $E$ is the number of electrodes and $T$ the number of timepoints.
The training objective is the optimal-transport conditional flow-matching (OT-CFM) loss~\citep{tong2024improving}.

\paragraph{Conditional flow matching.}
Flow matching~\citep{lipman2023flow} learns a velocity field $v_\theta(\vect{x},t):\R^{E\times T}\times[0,1]\rightarrow\R^{E\times T}$ whose induced flow transports an isotropic Gaussian prior $q_0$ into the data distribution $q_1$.
Following the two-endpoint formulation of~\citet{tong2024improving}, we specify a \emph{coupling} $\pi$, defined as a joint distribution over $(\vect{x}_0,\vect{x}_1)$ with noise marginal $q_0$ and data marginal $q_1$.
We then define a conditional probability path $p_t(\vect{x}\mid\vect{x}_0,\vect{x}_1)$ interpolating between $\vect{x}_0$ and $\vect{x}_1$, along with its conditional velocity field $u_t(\vect{x}\mid\vect{x}_0,\vect{x}_1)$.
 The conditional flow-matching loss,
\begin{equation}
  \label{eq:cfm}
  \mathcal{L}_{\mathrm{CFM}}(\theta)
  =\mathbb{E}_{t\sim\mathcal{U}[0,1],\;(\vect{x}_0,\vect{x}_1)\sim\pi,\;
    \vect{x}\sim p_t(\cdot\mid\vect{x}_0,\vect{x}_1)}
  \left\lVert v_\theta(\vect{x},t)
  -u_t(\vect{x}\mid\vect{x}_0,\vect{x}_1)\right\rVert^2
\end{equation}
shares the same gradient as the marginal flow-matching objective, meaning its minimizer is the true generative velocity.
We take the straight interpolant $\vect{x}_t=(1-t)\vect{x}_0+t\vect{x}_1$, for which $p_t(\cdot\mid\vect{x}_0,\vect{x}_1)$ is the Dirac delta at $\vect{x}_t$ and $u_t(\vect{x}\mid\vect{x}_0,\vect{x}_1)=\vect{x}_1-\vect{x}_0$.
Consequently, Eq.~\eqref{eq:cfm} reduces to regressing $v_\theta(\vect{x}_t,t)$ onto the displacement $\vect{x}_1-\vect{x}_0$.
The choice of coupling $\pi$ is the only remaining free parameter.
At inference, we sample by integrating $\dot{\vect{x}}_t=v_\theta(\vect{x}_t,t)$ from $\vect{x}_0\sim\mathcal{N}(\mathbf{0},\vect{I})$ at $t{=}0$ to $t{=}1$ using an off-the-shelf ordinary differential equation (ODE) solver.

\paragraph{Optimal transport coupling.}
Under the independent coupling $\pi=q_0\otimes q_1$, straight-line interpolants between random noise-data pairs cross in $\R^{E\times T}$, and the marginal velocity at crossing points must average incompatible conditional velocities.
This inflates gradient variance and curves the learned flow.
OT-CFM~\citep{tong2024improving} replaces $\pi$ with the minibatch optimal-transport plan under squared-Euclidean cost, pairing each noise sample with its nearest data sample inside the batch.
The resulting velocity field is straighter under the coupling, and few-step ODE sampling becomes accurate without distillation.
We solve the assignment exactly with the Hungarian algorithm and use the resulting permuted pairs in Eq.~\eqref{eq:cfm}.

\begin{figure}[t]
  \centering
  \includegraphics[width=0.95\linewidth]{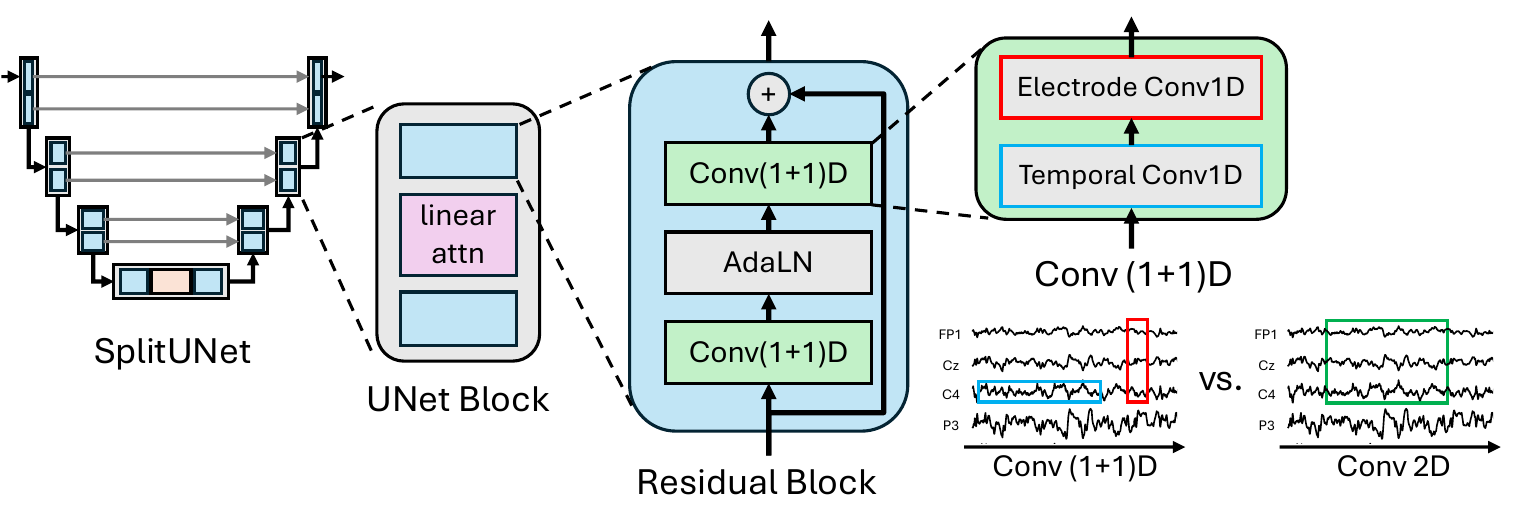}
  \vspace{-0.25cm}
  \caption{
  \textbf{Architecture of SplitUNet.}
(left) The encoder halves only the time axis at each of its three downsampling steps (electrode dimension preserved); a self-attention bottleneck mixes globally; the decoder mirrors the encoder with skip connections and time-only upsampling.
(center) Each UNet block contains two residual sub-blocks separated by a linear-attention layer~\citep{shen2021efficient}; each sub-block alternates Conv(1+1)D operators with AdaLN modulation conditioned on the flow time $t$.
(right) Conv(1+1)D factorizes a 2D convolution into a temporal Conv1D (blue, kernel $k_t$) followed by an electrode Conv1D (red, kernel $k_e$, sliding over electrodes in the standard 10--20 order); on an EEG snippet, this contrasts with the joint $(\text{electrode}, \text{time})$ patch of a Conv2D (green, kernel $k_e \times k_t$).
  }
  \vspace{-0.3cm}
  \label{fig:architecture}
\end{figure}
\subsection{SplitUNet: A Factorized Spatiotemporal Backbone for Multi-Channel Brain Signals}
\label{subsec:architecture}

Generative pretraining has converged on two dominant backbones: Transformers~\citep{ma2024sit,peebles2023scalable} and UNets~\citep{ho2020denoising,karras2022elucidating}.
However, Transformers~\citep{vaswani2017attention} require tokenizing the signal, thereby reintroducing the very discretization our continuous-time objective is designed to avoid.
A UNet-1D~\citep{ronneberger2015u} on the raw waveform removes that constraint, but, applied channel-wise, misses the inter-electrode correlations that carry most of the EEG's spatial information.
A UNet-2D over the $(\text{electrode}, \text{time})$ tensor restores electrode coupling, and a rectangular kernel ($k_t \gg k_e$) can, in principle, match the per-axis scale---time is locally autocorrelated and benefits from a wide temporal receptive field, while the electrode axis is short (only 19 channels in the 10--20 montage).

We propose \textit{SplitUNet}, a velocity-network backbone in which every $k_t \times k_e$ 2D convolution of a standard DDPM-style UNet~\citep{ho2020denoising} is replaced by the factorized (1+1)D operator
\begin{equation}
  \label{eq:oneonedconv}
\mathrm{conv}_{(1+1)\mathrm{D}}(\vect{z}) \;=\; \mathrm{conv}_{\text{electrode}}\ ( \sigma( \mathrm{conv}_{\text{time}}(\vect{z})\ )\ )
\end{equation}
where $\vect{z} \in \mathbb{R}^{C \times E \times T}$, the temporal convolution slides a kernel of size $k_t$ along the time axis (weights shared across electrodes), the electrode convolution slides a kernel of size $k_e$ along the electrode axis (weights shared across timepoints), and $\sigma$ is a non-linear activation.
This factorization, used in R(2+1)D for 3D video~\citep{tran2018closer}, has two advantages over the unfactorized 2D kernel: it interleaves a non-linearity between temporal and electrode mixing (doubling non-linear depth at no extra parameter cost), and it uses $k_t + k_e$ weights per filter rather than $k_t \cdot k_e$, redirecting the saved parameters into width or depth.
Empirically, it outperforms UNet-2D on every task (see Table~\ref{tab:ablation_backbone}), consistent with EEG's temporal dynamics and inter-electrode coupling being largely separable; we do not claim a 2D kernel is universally inferior.

Beyond the factorization, SplitUNet reflects two properties of multi-channel EEG: time is densely sampled, and the electrode axis is short, with each channel anatomically meaningful---compressing it would discard distinct brain-region information.
(i) \emph{Time-only compression.} The encoder halves only the temporal resolution after each of its first three stages and preserves the electrode dimension at $E$ throughout.
(ii) \emph{Attention placement.} Each resolution stage uses an efficient linear-attention block~\citep{shen2021efficient} for in-stage mixing, with a single full self-attention block at the bottleneck (where the time axis has been compressed by $8\times$) so that both axes are mixed once their dimensions are commensurate.
In both block types the $(E,T)$ feature map is flattened into a single sequence of length $E\cdot T$, so attention mixes over all electrode and time positions jointly and carries no axis-specific prior; the axis asymmetry is expressed by the convolutions and the downsampling schedule alone.
The decoder mirrors the encoder with skip connections and time-only upsampling.
Implementation details are in Appendix~\ref{supp:experiment_detail}, and the full architecture is illustrated in Figure~\ref{fig:architecture}.

\subsection{Pretraining and Feature Extraction}\label{subsec:pretraining}
We minimize the OT-CFM loss of Eq.~\eqref{eq:cfm} on the pretraining dataset (TUEG~\citep{obeid2016temple}).
Noise endpoints $\vect{x}_0$ are drawn from an isotropic Gaussian of matching shape, data endpoints $\vect{x}_1$ are sampled windows, and each minibatch is re-paired by an exact Hungarian algorithm before forming interpolants. Time $t$ is sampled uniformly on $[0,1]$.

For downstream transfer, we evaluate the velocity network on the input data $\vect{x}$ at $t=0$.
We extract features from the penultimate layer (immediately preceding the final output projection), yielding an activation tensor of shape $C \times E \times T$, where $C$ is the feature channel width, $E$ is the number of electrodes, and $T$ is the number of timepoints.
Applying global average pooling or flattening over both the electrode and time axes produces a fixed-size feature vector, which is then passed to a single linear classification head.
The complete finetuning protocol is described in Section~\ref{subsec:setup}.

\begin{table}[t!]
\caption{Overview of downstream tasks and datasets.
        }
\centering
\resizebox{0.95\linewidth}{!}{
    \small
    \begin{tabular}{ccccccc}
        \toprule
        \textbf{Task}&\textbf{Dataset}&\textbf{\# Channels}&\textbf{Duration}&\textbf{\# Samples}&\textbf{Rate}&\textbf{\# Classes} \\
        \midrule
        Mental Disorder&Mumtaz&19&5\,s&7,143&256\,Hz&2 \\
        Mental Stress&MAT&20&5\,s&1,707&500\,Hz&2 \\
        Seizure Detection & Siena & 29 & 10\,s & 51,307 & 512\,Hz & 2 \\
        \multirow{2}{*}{Sleep Staging}&ISRUC&6&30\,s&89,240&200\,Hz&5 \\
        & HMC & 4 & 30\,s & 137,243 & 256\,Hz & 5 \\
        Event Type&TUEV&16&5\,s&112,491&256\,Hz&6 \\
        Abnormal Detection&TUAB&16&10\,s&409,455&256\,Hz&2 \\
               Emotion Recognition&SEED-V & 62 & 1\,s & 117,744 & 1000\,Hz & 5 \\
        Motor Imagery & BCIC-IV-2a&22&4\,s&5,184&250\,Hz&4 \\
    \bottomrule
    \end{tabular}
    }
    \label{tab:downstream}
\end{table}

\section{Experiment}
\label{sec:exp}
\model replaces the masked-token recipe of existing EEG foundation models with continuous-time flow matching on the raw multi-channel waveform; we test this swap with two questions:
\textbf{(Q1) Downstream task performance:} Does the same pretrained backbone, with no task-specific architectural changes, transfer competitively to standard EEG classification benchmarks under matched finetuning, despite using no patches, no codebooks, and no masking (see Section~\ref{subsec:downstream})?
\textbf{(Q2) Clinical indistinguishability:} Does the same model also produce clinically realistic EEG, \ie, samples indistinguishable from real held-out TUEG to two board-certified neurologists (see Section~\ref{subsec:samples})?
Implementation details and hyperparameters are in Appendix~\ref{supp:experiment_detail}.

\subsection{Experimental Setup}\label{subsec:setup}

\paragraph{Pretraining corpus and preprocessing.}
We pretrain on the Temple University Hospital EEG Corpus~\citep{obeid2016temple} (TUEG): 69,652 recordings (27,062\,h) from 14,987 subjects across 26,846 sessions.
We follow established preprocessing pipelines~\citep{jiang2024large,ouahidi2025reve,wang2025cbramod} (band-pass and notch filtering, segmentation, amplitude thresholding, normalization; full settings in Appendix~\ref{supp:experiment_detail}) on the standard 19-channel 10--20 montage~\citep{klem1999ten} and its channel order, yielding 1,109,545 non-overlapping 30\,s segments ($\approx$\,9,000\,h).
 
From this pool we randomly sample a pretraining subset of 36,895 segments ($\approx$\,307\,h)---roughly 3.3\% of the preprocessed pool and 1\% of the TUEG hours---which is 1--2 orders of magnitude smaller than the corpora used by recent baselines (LaBraM: 2,500\,h~\citep{jiang2024large}; CBraMod: 9,000\,h~\citep{wang2025cbramod}; REVE: 60,000\,h (25,000 subjects across multiple corpora)~\citep{ouahidi2025reve}).

\paragraph{Pretraining.}
The velocity network is trained by minimizing the OT-CFM loss (see Eq.~\eqref{eq:cfm}) with an MSE regression target.
Each minibatch is re-paired by an exact optimal-transport assignment using the Hungarian algorithm before forming interpolants; $t$ is sampled uniformly on $[0,1]$.
We train for 70 epochs with batch size 4 and maintain an exponential moving average of the weights for sampling and downstream feature extraction. Optimizer settings are in Appendix~\ref{supp:experiment_detail}.

\paragraph{Downstream tasks.}
We evaluate on nine classification tasks spanning anomaly, seizure, and sleep-stage detection, motor imagery, and emotion/stress recognition (see Table~\ref{tab:downstream}).
We strictly follow published preprocessing pipelines, class-balance policies, and train/val/test splits: \citet{zhou2025csbrain} for Siena, HMC, and SEED-V; prior work~\citep{wang2025cbramod,yang2023biot,jiang2024large,ouahidi2025reve} for the remainder.
Dataset specifics are in Appendix~\ref{supp:datasets}.

\paragraph{Baselines.}
We compare \model against two groups: task-specific supervised architectures trained from scratch on each dataset, and pretrained EEG foundation models adapted by full-network finetuning.
The supervised group comprises EEGNet~\citep{lawhern2018eegnet}, EEGConformer~\citep{song2022eegconformer}, SPaRCNet~\citep{jing2023development}, ContraWR~\citep{yang2023self}, a CNN--Transformer hybrid~\citep{peh2022transformer}, FFCL~\citep{li2022motor}, and ST-Transformer~\citep{song2021transformer}.
The foundation-model group comprises BIOT~\citep{yang2023biot}, LaBraM-Base~\citep{jiang2024large}, CBraMod~\citep{wang2025cbramod}, REVE~\citep{ouahidi2025reve}, and CSBrain~\citep{zhou2025csbrain}.
All baseline numbers are taken directly from~\citet{ouahidi2025reve} and~\citet{zhou2025csbrain}; both papers score every baseline under finetuning on the same train/val/test splits we use, so the numbers in Table~\ref{tab:merged} are directly comparable to ours.

\paragraph{Finetuning protocol.}
We finetune the pretrained \model end-to-end on each downstream task with cross-entropy. The backbone is conditioned on a fixed timestep $t{=}0$; its output features are passed through a single linear head.
The UNet-1D (joint) variant has an electrode-count-dependent first convolution, which we reinitialize per dataset before finetuning; the other backbones are electrode-count-invariant.
All downstream results are averaged over the five highest-scoring seeds, with each run reported at its best epoch on the test split.
Please refer to Appendix~\ref{supp:experiment_detail} for more details.

\begin{table}[t!]
    \centering
    \caption{
    Balanced accuracy ($\pm$ std) across nine downstream classification tasks. \model reaches the highest suite-averaged balanced accuracy ($0.754$), sets a new state of the art on $7$ of $9$ tasks, is within $1.5$\,pp on TUAB, and achieves second-best performance on BCIC-IV-2a (motor imagery on healthy subjects).
Bold and underline denote the best and the second best performance in each task.
    }
    \resizebox{\textwidth}{!}{%
    \begin{tabular}{lccccc}
        \toprule
         \textbf{Methods}&  Mumtaz&  MAT&  Siena&  ISRUC& HMC\\
         \midrule
         EEGNet&  0.923 $\pm$ 0.010&  0.677 $\pm$ 0.012&  0.749 $\pm$ 0.052&  0.715 $\pm$ 0.012& 0.653 $\pm$ 0.012\\
         EEGConformer&  0.931 $\pm$ 0.012&  0.680 $\pm$ 0.012&  0.756 $\pm$ 0.021&  0.740 $\pm$ 0.013& 0.715 $\pm$ 0.009\\
         SPaRCNet&  0.932 $\pm$ 0.009&  0.688 $\pm$ 0.011&  0.657 $\pm$ 0.038&  0.749 $\pm$ 0.007& 0.476 $\pm$ 0.111\\
         ContraWR&  0.919 $\pm$ 0.011&  0.663 $\pm$ 0.010&  0.655 $\pm$ 0.031&  0.740 $\pm$ 0.013& 0.424 $\pm$ 0.054\\
         CNN-Transformer&  0.930 $\pm$ 0.007&  0.678 $\pm$ 0.027&  0.698 $\pm$ 0.056&  0.736 $\pm$ 0.009& 0.657 $\pm$ 0.014\\
         FFCL&  0.931 $\pm$ 0.004&  0.680 $\pm$ 0.014&  0.662 $\pm$ 0.039&  0.728 $\pm$ 0.018& 0.443 $\pm$ 0.070\\
         ST-Transformer&  0.913 $\pm$ 0.010&  0.663 $\pm$ 0.017&  0.753 $\pm$ 0.038&  0.738 $\pm$ 0.021& 0.256 $\pm$ 0.014\\
         \midrule
         BIOT&  0.936 $\pm$ 0.005&  0.688 $\pm$ 0.019&  0.735 $\pm$ 0.067&  0.753 $\pm$ 0.012& 0.686 $\pm$ 0.004\\
         LaBraM-Base&  0.941 $\pm$ 0.008&  0.691 $\pm$ 0.013&  0.708 $\pm$ 0.033&  0.763 $\pm$ 0.010& 0.728 $\pm$ 0.010\\
         CBraMod&  0.956 $\pm$ 0.006&  0.726 $\pm$ 0.013&  0.732 $\pm$ 0.065&  0.786 $\pm$ 0.011& 0.727 $\pm$ 0.004\\
         REVE&  \underline{0.964 $\pm$ 0.010} &  \underline{0.766 $\pm$ 0.035}& 0.740 $\pm$ 0.007 &  0.782 $\pm$ 0.008& \underline{0.740 $\pm$ 0.007}\\
         CSBrain&  0.964 $\pm$ 0.015&  0.756 $\pm$ 0.011&  \underline{0.766 $\pm$ 0.047}&  \underline{0.792 $\pm$ 0.003}& 0.735 $\pm$ 0.005\\
          \midrule
        \model (ours) & \textbf{\mumtazBalAccSweep} & \textbf{\matBalAccSweep} & \textbf{\sienaBalAcc} & \textbf{\isrucBalAcc} & \textbf{\hmcBalAcc} \\
          \midrule
         & TUEV&  TUAB& BCIC-IV-2a& SEED-V & \cellcolor[HTML]{FAF0F0} Average\\
         \midrule
    EEGNet & 0.388 $\pm$ 0.014 & 0.764 $\pm$ 0.004 & 0.448 $\pm$ 0.009 & 0.296 $\pm$ 0.010 & \cellcolor[HTML]{FAF0F0}0.624 $\pm$ 0.020\\
    EEGConformer & 0.407 $\pm$ 0.016 & 0.776 $\pm$ 0.005 & 0.470 $\pm$ 0.011 & 0.354 $\pm$ 0.011 & \cellcolor[HTML]{FAF0F0}0.648 $\pm$ 0.013\\
    SPaRCNet & 0.416 $\pm$ 0.026 & 0.790 $\pm$ 0.002 & 0.464 $\pm$ 0.012 & 0.295 $\pm$ 0.008 & \cellcolor[HTML]{FAF0F0}0.607 $\pm$ 0.041\\
    ContraWR & 0.438 $\pm$ 0.035 & 0.775 $\pm$ 0.004 & 0.468 $\pm$ 0.013 & 0.355 $\pm$ 0.011 & \cellcolor[HTML]{FAF0F0}0.604 $\pm$ 0.025\\
    CNN-Transformer & 0.409 $\pm$ 0.016 & 0.778 $\pm$ 0.002 & 0.460 $\pm$ 0.011 & 0.368 $\pm$ 0.008 & \cellcolor[HTML]{FAF0F0}0.635 $\pm$ 0.023\\
    FFCL & 0.398 $\pm$ 0.010 & 0.785 $\pm$ 0.004 &  0.447 $\pm$ 0.014 & 0.364 $\pm$ 0.009 & \cellcolor[HTML]{FAF0F0}0.604 $\pm$ 0.029\\
    ST-Transformer & 0.398 $\pm$ 0.023 & 0.797 $\pm$ 0.002  & 0.458 $\pm$ 0.015 & 0.305 $\pm$ 0.007 & \cellcolor[HTML]{FAF0F0}0.587 $\pm$ 0.019\\
         \midrule
    BIOT & 0.528 $\pm$ 0.022 & 0.796 $\pm$ 0.006 & 0.475 $\pm$ 0.009 & 0.384 $\pm$ 0.019 & \cellcolor[HTML]{FAF0F0}0.664 $\pm$ 0.026\\
    LaBraM-Base & 0.641 $\pm$ 0.006 & 0.814 $\pm$ 0.002 &  0.487 $\pm$ 0.009 & 0.398 $\pm$ 0.014 & \cellcolor[HTML]{FAF0F0}0.686 $\pm$ 0.014\\
    CBraMod & 0.667 $\pm$ 0.011 & \underline{0.829 $\pm$ 0.002} &  0.514 $\pm$ 0.007 & 0.409 $\pm$ 0.010 & \cellcolor[HTML]{FAF0F0}0.705 $\pm$ 0.023\\
    REVE & 0.676 $\pm$ 0.023 & \textbf{0.832 $\pm$ 0.001} & \textbf{0.640 $\pm$ 0.009} & 0.405 $\pm$ 0.002 & \cellcolor[HTML]{FAF0F0}\underline{0.727 $\pm$ 0.015}\\
    CSBrain & \underline{0.690 $\pm$ 0.006} & 0.817 $\pm$ 0.004 & 0.566 $\pm$ 0.007 & \underline{0.420 $\pm$ 0.003} & \cellcolor[HTML]{FAF0F0}0.723 $\pm$ 0.017 \\
         \midrule
         \model (ours) & \textbf{\tuevBalAcc} & \tuabBalAcc & \underline{\bcicivaBalAccSweep} & \textbf{\seedvBalAcc} & \cellcolor[HTML]{FAF0F0}\textbf{0.754 $\pm$ 0.010}  \\
         \bottomrule
    \end{tabular}
    }
    \label{tab:merged}
\end{table}

\subsection{Downstream Classification}\label{subsec:downstream}

We address \textbf{Q1} (downstream transfer) by finetuning the pretrained \model backbone end-to-end on each of the nine tasks under the protocol of Section~\ref{subsec:setup}.
Table~\ref{tab:merged} reports balanced accuracy per task; per-task breakdowns with task-appropriate companion metrics (Cohen's $\kappa$, Weighted F1, AUC-PR, AUROC) are provided in Appendix~\ref{supp:per_task}.
\model achieves a new state of the art on 7 of the 9 tasks (Mumtaz, MAT, Siena, ISRUC, HMC, TUEV, and SEED-V), and trails the leading model by less than $1.5$\,pp on TUAB. 
However, a performance gap remains on the motor imagery task, BCIC-IV-2a, where \model achieves the second-best performance but lags behind REVE ($0.570$ vs.\ $0.640$). 
This gap is largely attributable to differences in the pretraining corpora.
REVE explicitly leverages four motor imagery datasets~\citep{cho2017eeg,ofner2017upper,liu2024eeg,zhou2016fully} during pretraining, making its downstream evaluation on this task effectively in-domain.
In contrast, \model is pretrained exclusively on TUEG---a clinical corpus of resting-state and pathological EEG. 
Thus, while REVE enjoys target-domain familiarity, \model still achieves highly competitive cross-domain transfer, successfully adapting to motor imagery signals during finetuning despite never encountering them during pretraining.

\subsection{Clinical Indistinguishability of Generated Samples}\label{subsec:samples}
\begin{figure}[t!]
  \centering
  \includegraphics[width=\linewidth]{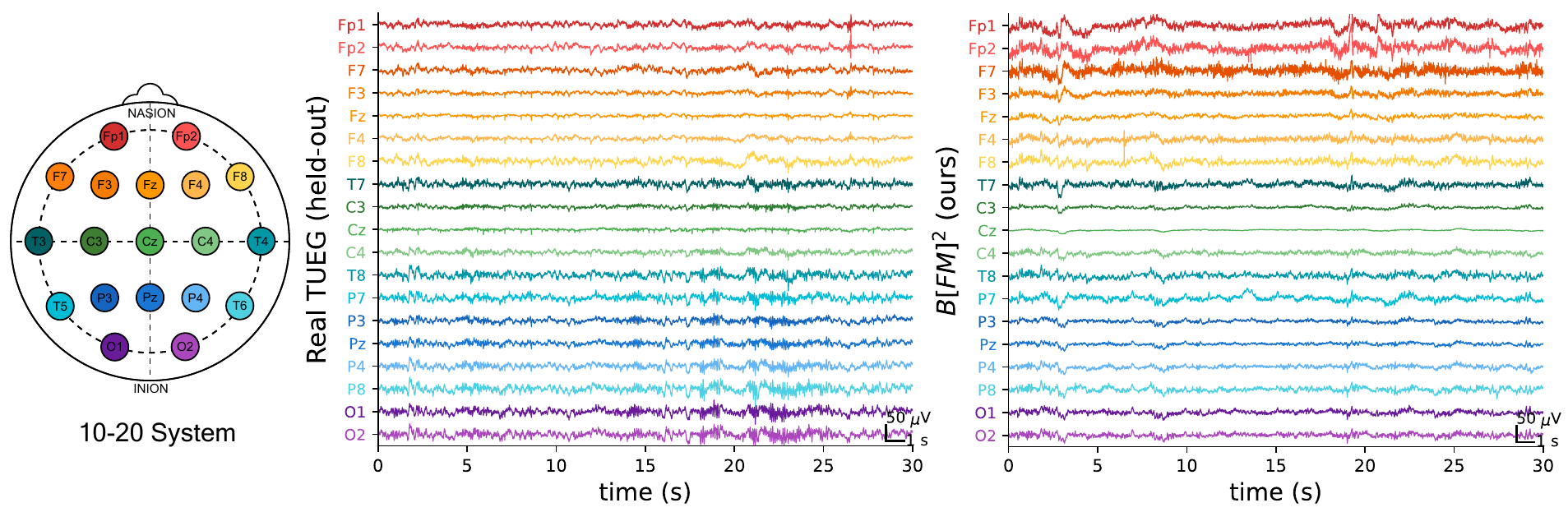}
\caption{\textbf{Real vs.\ \model EEG segments.}
A 30-second held-out TUEG segment (middle) and an unconditional \model sample (right), displayed in the 19-channel referential 10--20 montage (left; electrode layout from~\citet{ferrell2020temple}).
Both panels exhibit both the spatial and temporal coherence characteristic of clinical EEG. For example, notice how adjacent electrodes in the same brain region (e.g., the frontal red traces or the occipital purple traces) share rhythmically aligned activity and correlated amplitude envelopes, rather than exhibiting independent, per-channel noise.
The two sources are visually difficult to distinguish at a glance, consistent with the blinded neurologist evaluation in Table~\ref{tab:neurologist}.
Note that the right-hand traces are an \emph{independent} unconditional sample drawn from Gaussian noise, not a reconstruction of a real segment.
Calibration: $1$\,s, $50\,\mu$V.
  }
  \label{fig:samples}
  \vspace{-0.3cm}
\end{figure}

To address \textbf{Q2} (clinical indistinguishability), we generate unconditional samples using the same pretrained \model from Section~\ref{subsec:downstream}.
As shown in the example traces in Figures~\ref{fig:samples} and~\ref{fig:supp_samples}), the generated signals are visually compelling.
In a blinded reading, two board-certified neurologists could not reliably distinguish them from real, held-out TUEG segments (see Table~\ref{tab:neurologist}).
Section~\ref{supp:qualitative} additionally showcases physiologically diverse patterns---sharp-wave transients, slow-wave activity, and posterior $\alpha$ rhythm---identified in generated samples (see Figure~\ref{fig:gen_showcase}).
The two neurologists each evaluated 50 randomly interleaved 30-second segments (25 real held-out TUEG and 25 unconditional \model samples). 
Segments were presented in the longitudinal bipolar 10--20 montage with the standard clinical display settings under which they habitually read EEG.
The neurologists scored each segment on a 5-point Likert scale of perceived realness ($1 = \text{definitely real}$, $3 = \text{unsure}$, $5 = \text{definitely fake}$).

\begin{table}[t!]
\centering
\caption{Blinded neurologist rating of \model samples.
Two board-certified neurologists evaluated 50 randomly interleaved 30-second windows each (25 real held-out TUEG and 25 \model samples).
Realness was scored on a 5-point Likert scale ($1 = \text{definitely real}$, $5 = \text{definitely fake}$).
Readers failed to distinguish synthetic samples from real segments, rating \model outputs as slightly more realistic on average.
}
 \vspace{0.1cm}
 \label{tab:neurologist}
\resizebox{0.9\linewidth}{!}{%
\begin{tabular}{lcccc|cc}
\toprule
 \multirow{2}{*}{Reader} & \multicolumn{2}{c}{Mean realness ($\downarrow$ for real, $\uparrow$ for fake)} & \multicolumn{2}{c|}{Mann--Whitney $U$} & \multicolumn{2}{c}{Agreement} \\
\cmidrule(lr){2-7}
& Real Data & \model & U & $p$ & $\rho$ & Cohen's $\kappa$ \\
\midrule
R1    & 2.96 $\pm$ 1.02 & 2.60 $\pm$ 1.04 & 251.5 & 0.204 & \multirow{2}{*}{0.052} & \multirow{2}{*}{$-0.096$} \\
R2     &3.08 $\pm$ 1.12 & 2.32 $\pm$ 0.80 & 188.5 &0.011 &                        \\
\midrule
Pooled &3.02 $\pm$ 1.06& 2.46 $\pm$ 0.93 & 876.0 & 0.007 & - & -                   \\
\bottomrule
\end{tabular}
}
\vspace{-0.3cm}
\end{table}

Pooled across readers, the mean realness score was $3.02 \pm 1.06$ for real segments versus $2.46 \pm 0.93$ for \model samples (Mann--Whitney $U = 876.0$, $p = 0.007$).
This statistically significant gap points in an unexpected direction: \model samples were rated slightly \textit{more} real-looking than the actual held-out EEG segments.
Furthermore, at the segment level, the raters failed to agree on the origin of the samples (Spearman $\rho = 0.052$, Cohen's $\kappa = -0.096$).
Neither reader could reliably classify an individual segment as real or synthetic, indicating that the perceptual cues they relied upon were largely uncorrelated.
The full reading protocol and clinical interface are detailed in Appendix~\ref{supp:neurologist_protocol}.

\subsection{Recognizable Brain-State Patterns from Unconditional Generation}\label{supp:qualitative}
\begin{figure}[t!]
  \centering
  \includegraphics[width=\linewidth]{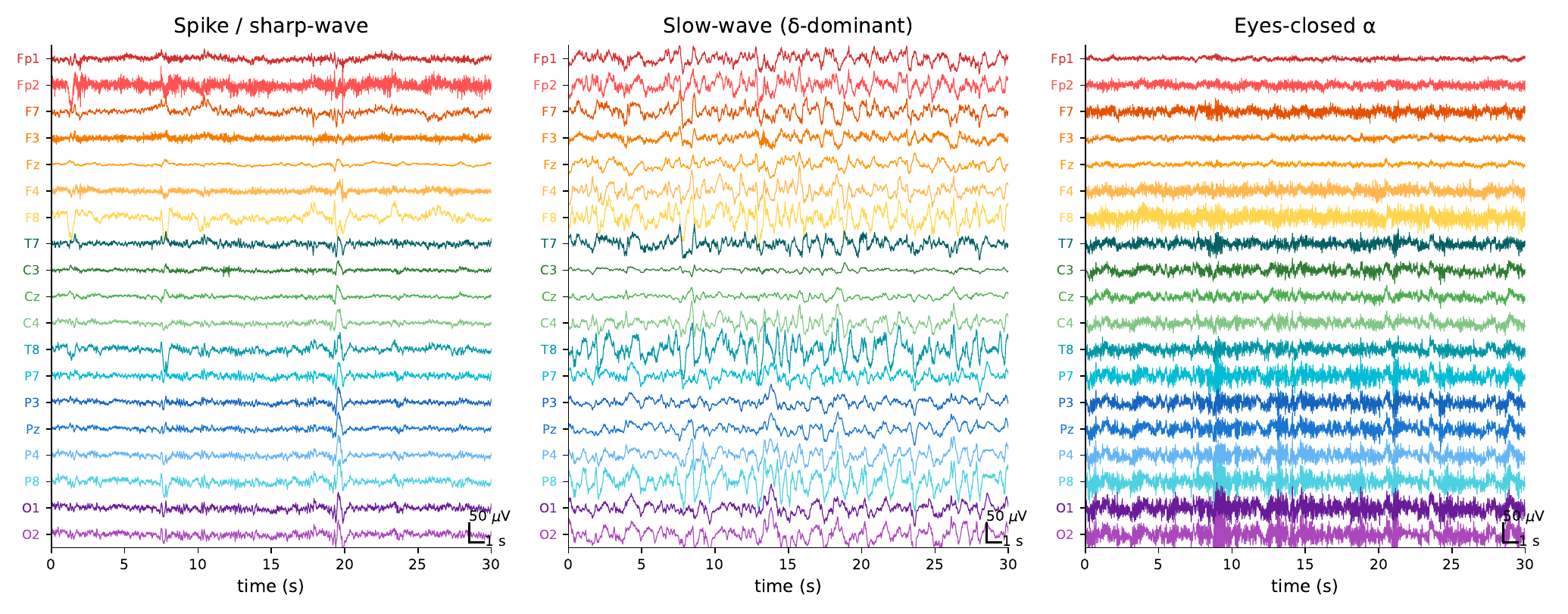}
  \vspace{-0.3cm}
  \caption{\textbf{\model generates physiologically diverse EEG patterns.}
  Three \model samples illustrating distinct brain-state patterns, curated from the unconditional generation pool by simple heuristics and displayed in the canonical 10--20 montage.
  (left) a sharp-wave / spike-like transient at $\sim$20\,s.
  (center) $\delta$-dominant slow-wave activity across all electrodes, characteristic of slow-wave sleep.
  (right) eyes-closed posterior $\alpha$ rhythm, with stronger 8--13\,Hz power over O1/O2.
Calibration bar: $1$\,s and $50\,\mu$V. 
  }
  \label{fig:gen_showcase}
  \vspace{-0.3cm}
\end{figure}

\model captures recognizable brain-state structure from a fully unconditional objective.
Mining its generation pool with simple band-power heuristics~(see Appendix~\ref{supp:heuristics}), we recover three canonical EEG patterns: a sharp-wave transient, $\delta$-dominant slow-wave activity, and posterior $\alpha$ rhythm (see Figure~\ref{fig:gen_showcase}). Each appears alongside spatially coherent activity across adjacent electrodes (see Figure~\ref{fig:samples}), as expected of clinical recordings.
This kind of qualitative inspection is itself uncommon in the EEG foundation-model literature: tokenized masked-token backbones~\citep{jiang2024large,wang2025cbramod,ouahidi2025reve,yang2023biot} are not designed to generate novel waveforms in the input space and so cannot be probed for what they have learned about brain dynamics.
That \model reveals interpretable structure without any class or pattern labels also points to a natural extension---class- and subject-conditional synthesis to augment scarce, long-tailed clinical EEG corpora (see Section~\ref{sec:discussion}).

\subsection{Ablation Studies}\label{subsec:ablation}

We conduct ablation studies on the network architecture (see Tables~\ref{tab:sample-quality-arch} and~\ref{tab:ablation_backbone}), the pretraining objective (see Table~\ref{tab:ablation_objective}), and pretraining itself (see Table~\ref{tab:ablation_pretrain}) to assess the contribution of each component.

\begin{table}[t]
  \centering
  \caption{Sample quality of \model with four velocity-network backbones against held-out real TUEG segments ($n{=}5{,}000$, mean $\pm$ std).
  UNet-1D (indep.) processes each electrode as a separate time series, so no information passes between electrodes.
  UNet-1D (joint) takes all electrodes as the input channels of a temporal UNet, so electrodes are combined only in its first convolution.
  UNet-2D applies 2D convolutions over electrodes and time at every layer, and SplitUNet replaces each 2D convolution with a temporal convolution followed by an electrode convolution.
  Real-vs-Real compares two disjoint sets of real segments and gives the distance expected from sampling alone.
  Bold marks the best generator per column, excluding Real-vs-Real.
  PSNR and SSIM are computed without correspondence between generated and real segments and are reported for completeness.}
  \vspace{0.1cm}
  \resizebox{0.7\linewidth}{!}{
  \begin{tabular}{lccc}
  \toprule
  Backbone & Band-power $W_1$ $\downarrow$ & PSNR $\uparrow$ & SSIM~$\uparrow$ \\
  \midrule
  Real-vs-Real & 0.156 $\pm$ 0.028 & 16.32 $\pm$ 0.26 & 0.021 $\pm$ 0.003 \\
  \midrule
  UNet-1D (indep.) & 1.903 $\pm$ 0.098 & 6.74 $\pm$ 0.07 & 0.001 $\pm$ 0.000 \\
  UNet-1D (joint) & 0.256 $\pm$ 0.002 & 16.88 $\pm$ 0.03 & 0.025 $\pm$ 0.001 \\
  UNet-2D & 0.248 $\pm$ 0.001 & 17.14 $\pm$ 0.01 & 0.024 $\pm$ 0.001 \\
  SplitUNet (ours) & \textbf{0.195 $\pm$ 0.002} & \textbf{17.16 $\pm$ 0.02} & \textbf{0.026 $\pm$ 0.001} \\
  \bottomrule
  \end{tabular}
  }
  \label{tab:sample-quality-arch}
\end{table}

\begin{table}[t]
  \centering
  \caption{
   Balanced accuracy ($\pm$ std) of \model with four velocity-network backbones, each pretrained with OT-CFM on the same $307$\,h set and finetuned end-to-end on the nine downstream tasks.
   UNet-1D (indep.) processes each electrode as a separate time series, so no information passes between electrodes.
  UNet-1D (joint) takes all electrodes as the input channels of a temporal UNet.
   Bold denotes the best performance.
  }
  \vspace{0.1cm}
  \small
  \begin{tabular}{lcccc}
  \toprule
  \textbf{Dataset} & UNet-1D (indep.) & UNet-1D (joint) & UNet-2D & SplitUNet (ours) \\
  \midrule
  Mumtaz     & \OneDIdentitymumtazBalAcc & \OneDmumtazBalAcc & \TwoDmumtazBalAcc & \textbf{\mumtazBalAccSweep} \\
  MAT        & \OneDIdentitymatBalAccSweep & \OneDmatBalAccSweep & \TwoDmatBalAccSweep & \textbf{\matBalAccSweep} \\
  Siena      & \OneDIdentitysienaBalAcc & \OneDsienaBalAcc & \TwoDsienaBalAcc & \textbf{\sienaBalAcc} \\
  ISRUC      & \OneDIdentityisrucBalAcc & \OneDisrucBalAcc & \TwoDisrucBalAcc & \textbf{\isrucBalAcc} \\
  HMC        & \OneDIdentityhmcBalAcc & \OneDhmcBalAcc & \TwoDhmcBalAcc & \textbf{\hmcBalAcc} \\
  TUEV       & \OneDIdentitytuevBalAcc & \OneDtuevBalAcc & \TwoDtuevBalAcc & \textbf{\tuevBalAcc} \\
  TUAB       & \OneDIdentitytuabBalAcc & \textbf{\OneDtuabBalAcc} & \TwoDtuabBalAcc & \tuabBalAcc \\
  BCIC-IV-2a & \OneDIdentitybcicivaBalAccSweep & \OneDbcicivaBalAccSweep & \TwoDbcicivaBalAccSweep & \textbf{\bcicivaBalAccSweep} \\
  SEED-V     & \OneDIdentityseedvBalAccSweep & \OneDseedvBalAccSweep & \TwoDseedvBalAccSweep & \textbf{\seedvBalAcc} \\
  \midrule
  \cellcolor[HTML]{FAF0F0}Average & \cellcolor[HTML]{FAF0F0}0.674 $\pm$ 0.020 & \cellcolor[HTML]{FAF0F0}0.682 $\pm$ 0.016 & \cellcolor[HTML]{FAF0F0}0.686 $\pm$ 0.025 & \cellcolor[HTML]{FAF0F0}\textbf{0.754 $\pm$ 0.010} \\
  \bottomrule
  \end{tabular}
  \label{tab:ablation_backbone}
\end{table}

\begin{table}[t]
  \centering
  \caption{Balanced accuracy ($\pm$ std) of SplitUNet pretrained with pointwise masked reconstruction, patchwise ($1$\,s) masked reconstruction, or OT-CFM on the same pretraining data.
Bold denotes the best performance.}
  \vspace{0.1cm}
  \small
  \begin{tabular}{lccc}
  \toprule
  \textbf{Dataset} & Masked (pointwise) & Masked (patchwise) & OT-CFM (ours) \\
  \midrule
  Mumtaz     & 0.878 $\pm$ 0.009 & 0.797 $\pm$ 0.033 & \textbf{\mumtazBalAccSweep} \\
  MAT        & 0.704 $\pm$ 0.096 & 0.732 $\pm$ 0.085 & \textbf{\matBalAccSweep} \\
  Siena      & 0.764 $\pm$ 0.021 & 0.750 $\pm$ 0.023 & \textbf{\sienaBalAcc} \\
  ISRUC      & 0.795 $\pm$ 0.004 & 0.794 $\pm$ 0.006 & \textbf{\isrucBalAcc} \\
  HMC        & 0.755 $\pm$ 0.006 & 0.755 $\pm$ 0.007 & \textbf{\hmcBalAcc} \\
  TUEV       & 0.669 $\pm$ 0.025 & 0.637 $\pm$ 0.030 & \textbf{\tuevBalAcc} \\
  TUAB       & 0.822 $\pm$ 0.006 & \textbf{0.823 $\pm$ 0.005} & \tuabBalAcc \\
  BCIC-IV-2a & 0.522 $\pm$ 0.030 & 0.557 $\pm$ 0.016 & \textbf{\bcicivaBalAccSweep} \\
  SEED-V     & 0.484 $\pm$ 0.014 & 0.457 $\pm$ 0.034 & \textbf{\seedvBalAcc} \\
  \midrule
  \cellcolor[HTML]{FAF0F0}Average & \cellcolor[HTML]{FAF0F0}0.710 $\pm$ 0.036 & \cellcolor[HTML]{FAF0F0}0.700 $\pm$ 0.035 & \cellcolor[HTML]{FAF0F0}\textbf{0.754 $\pm$ 0.010} \\
  \bottomrule
  \end{tabular}
  \label{tab:ablation_objective}
\end{table}

\begin{table}[t]
  \centering
  \caption{Balanced accuracy ($\pm$ std) of CBraMod and \model trained from scratch on each downstream dataset or pretrained on our $307$\,h set.
  The CBraMod $9{,}000$\,h column is from~\citet{wang2025cbramod}.
  Bold denotes the best performance.
  }
  \vspace{0.1cm}
  \resizebox{0.9\linewidth}{!}{%
  \begin{tabular}{lccccc}
  \toprule
  & \multicolumn{3}{c}{CBraMod} & \multicolumn{2}{c}{\model} \\
  \cmidrule(lr){2-4}\cmidrule(lr){5-6}
  \textbf{Dataset} & Scratch & $307$\,h & $9{,}000$\,h & Scratch & $307$\,h (ours) \\
  \midrule
  Mumtaz     & 0.895 $\pm$ 0.009 & 0.909 $\pm$ 0.013 & 0.956 $\pm$ 0.006 & 0.863 $\pm$ 0.042 & \textbf{\mumtazBalAccSweep} \\
  MAT        & 0.623 $\pm$ 0.045 & 0.623 $\pm$ 0.046 & 0.726 $\pm$ 0.013 & 0.669 $\pm$ 0.059 & \textbf{\matBalAccSweep} \\
  Siena      & 0.814 $\pm$ 0.014 & \textbf{0.816 $\pm$ 0.021} & 0.732 $\pm$ 0.065 & 0.595 $\pm$ 0.140 & \sienaBalAcc \\
  ISRUC      & 0.776 $\pm$ 0.006 & 0.775 $\pm$ 0.006 & 0.786 $\pm$ 0.011 & 0.762 $\pm$ 0.063 & \textbf{\isrucBalAcc} \\
  HMC        & 0.720 $\pm$ 0.009 & 0.720 $\pm$ 0.009 & 0.727 $\pm$ 0.004 & 0.736 $\pm$ 0.012 & \textbf{\hmcBalAcc} \\
  TUEV       & 0.678 $\pm$ 0.017 & 0.648 $\pm$ 0.011 & 0.667 $\pm$ 0.011 & 0.593 $\pm$ 0.030 & \textbf{\tuevBalAcc} \\
  TUAB       & 0.801 $\pm$ 0.011 & 0.790 $\pm$ 0.011 & \textbf{0.829 $\pm$ 0.002} & 0.824 $\pm$ 0.005 & \tuabBalAcc \\
  BCIC-IV-2a & 0.403 $\pm$ 0.020 & 0.386 $\pm$ 0.029 & 0.514 $\pm$ 0.007 & 0.253 $\pm$ 0.003 & \textbf{\bcicivaBalAccSweep} \\
  SEED-V     & 0.334 $\pm$ 0.013 & 0.404 $\pm$ 0.005 & 0.409 $\pm$ 0.010 & 0.419 $\pm$ 0.023 & \textbf{\seedvBalAcc} \\
  \midrule
  \cellcolor[HTML]{FAF0F0}Average & \cellcolor[HTML]{FAF0F0}0.672 $\pm$ 0.019 & \cellcolor[HTML]{FAF0F0}0.675 $\pm$ 0.021 & \cellcolor[HTML]{FAF0F0}0.705 $\pm$ 0.023 & \cellcolor[HTML]{FAF0F0}0.635 $\pm$ 0.058 & \cellcolor[HTML]{FAF0F0}\textbf{0.754 $\pm$ 0.010} \\
  \bottomrule
  \end{tabular}
  }
  \label{tab:ablation_pretrain}
\end{table}

\paragraph{Network architecture.}
We replace SplitUNet with three backbones that differ in how they couple electrodes.
\textbf{UNet-1D (indep.)} applies every temporal convolution to each electrode separately, so no information crosses electrodes anywhere in the network.
\textbf{UNet-1D (joint)} treats the $E$ electrodes as the input channels of the UNet.
Its first convolution is reinitialized at finetuning to match each dataset's electrode count.
\textbf{UNet-2D} convolves jointly over (electrode, time) at every layer.
For generation, we draw $5{,}000$ unconditional samples from each backbone and measure the Wasserstein-1 ($W_1$) distance between their log-band-power distributions and those of held-out TUEG segments (see Table~\ref{tab:sample-quality-arch}).
Electrode coupling is necessary for generation, as UNet-1D (indep.) samples lie at $W_1{=}1.903$ against a real-vs-real floor of $0.156$.
SplitUNet attains the lowest distance at $0.195$, compared with $0.248$ for UNet-2D and $0.256$ for UNet-1D (joint).

In downstream transfer (see Table~\ref{tab:ablation_backbone}), coupling electrodes without factorization raises the average balanced accuracy by at most $0.012$ over UNet-1D (indep.), from $0.674$ to $0.682$ for UNet-1D (joint) and $0.686$ for UNet-2D.
UNet-1D (indep.) even exceeds both coupled backbones on TUEV and SEED-V.
SplitUNet raises the average to $0.754$ and leads all three backbones on every task except TUAB, where UNet-1D (joint) is $0.007$ higher.
The largest margins over the best alternative appear on MAT ($+0.125$), BCIC-IV-2a ($+0.102$), and Mumtaz ($+0.091$).

\paragraph{Pretraining objective.}
We isolate the contribution of the flow-matching objective by replacing OT-CFM pretraining with masked reconstruction of either individual samples (\emph{pointwise}) or contiguous $1$\,s blocks (\emph{patchwise}), keeping SplitUNet and the pretraining data fixed (see Table~\ref{tab:ablation_objective}).
The pointwise variant masks $75\%$ of the $(E, T)$ samples uniformly at random and regresses the masked entries with an MSE loss on the waveform.
This applies the MAE formulation and masking ratio~\citep{he2022masked} at the sample level rather than the patch level, so no tokenization is introduced.
The patchwise variant instead masks $50\%$ of each recording, following the masking ratio of CBraMod~\citep{wang2025cbramod}, in contiguous blocks of $200$ samples ($1$\,s at $200$\,Hz) spanning all electrodes.
Masking entire blocks prevents the model from reconstructing a masked sample by interpolating from its immediate temporal neighbors.
Each pretrained backbone is finetuned with the protocol of Section~\ref{subsec:setup}.
OT-CFM raises the average balanced accuracy from $0.710$ (pointwise) and $0.700$ (patchwise) to $0.754$ and leads on every task except TUAB, where patchwise masking is $0.004$ higher.

\paragraph{Pretraining scale and initialization.}
Because the baselines are pretrained on $9{,}000$--$60{,}000$\,h, comparisons with them conflate the pretraining method with data volume.
To separate the two, we pretrain CBraMod~\citep{wang2025cbramod} on our $307$\,h pretraining set ($36{,}895$ segments) with its official codebase.
At this scale, CBraMod's masked pretraining barely improves on training from scratch, raising the average balanced accuracy from $0.672$ to $0.675$ (see Table~\ref{tab:ablation_pretrain}).
Pretraining \model on the same data raises its average from $0.635$ to $0.754$, even though \model trails CBraMod when both are trained from scratch.
\model thus exceeds the matched-scale CBraMod by $0.079$ on average and outperforms it on every task except Siena.
It also surpasses the published CBraMod pretrained on $9{,}000$\,h ($0.705$) with about one-thirtieth of the pretraining data.


\section{Conclusion}
\label{sec:discussion}

We propose \model, a brain foundation model that pretrains directly on the continuous multi-channel EEG waveform via flow matching. 
However, pretraining on EEG data presents a challenge of integrating two distinctive dimensions: a high-resolution temporal dimension and a fixed, low-resolution electrode dimension.
We introduced SplitUNet, a velocity-network backbone that factorizes each 2D spatiotemporal convolution into a 1D temporal and a 1D electrode convolution and preserves the electrode dimension at every layer.
\model establishes a new state of the art on $7$ of $9$ standard EEG downstream tasks, outperforming the standard practice of discretizing EEG into patches~\citep{ouahidi2025reve,zhou2025csbrain} or tokens~\citep{jiang2024large}.
Crucially, we achieve this using only $36{,}895$ segments---roughly $3.3$\,\% of the pretraining pool---while unconditionally generating samples that two board-certified neurologists cannot reliably distinguish from real clinical data.
Our paper has a limitation that \model is pretrained only on $307$\,h from a single clinical corpus with a fixed $19$-channel montage, so whether its advantage persists at the $9{,}000$--$60{,}000$\,h scale of existing models remains untested.
Beyond EEG, SplitUNet may extend naturally to other multi-channel time series with structured sensor arrays such as MEG and ECoG.
Class- and subject-conditional extensions also open the potential to synthesize rare clinical cases, accelerating clinical neuroscience research.

\section*{Acknowledgment}
We thank the MIT Office of Research Computing and Data (ORCD) for support through ORCD Seed
Fund Grants, which provided access to GPUs and additional funding support.
We also thank NVIDIA and AMD for supporting GPUs.
This work was supported by the Flemish Government under METHUSALEM grant METH/26/003 (Methusalem-BioMedAI-Explainable and generative AI for accelerating biomedical discoveries: from genome to function) and VLAIO project HBC.2025.0582 (MediSynth).

{
\small
\bibliographystyle{plainnat} 
\bibliography{egbib}
}

\appendix
\section*{Appendix}
\section{Related Work}
\label{sec:related}

\subsection{Brain Foundation Model}\label{subsec:related_bfm}

EEG foundation models aim to learn a general representation from large-scale brain signals (\ie, EEG) to have better performance on diverse downstream tasks, as opposed to task-specific methods~\citep{lawhern2018eegnet,song2022eegconformer}.
Training approaches for current models are generally classified into two categories: contrastive learning~\citep{chen2020simple} and masked reconstruction~\citep{he2022masked}.
Early contrastive approaches, such as BENDR~\citep{kostas2021bendr} and Brant~\citep{zhang2023brant}, demonstrated the feasibility of joint representation learning for physiological signals.
Concurrently, BIOT~\citep{yang2023biot} uses a Transformer~\citep{vaswani2017attention} with masked reconstruction~\citep{he2022masked}, which became a standard technique for subsequent methods~\citep{jiang2024large,ouahidi2025reve,wang2024eegpt,wang2025cbramod,zhou2025csbrain}.
Recent works focus on incorporating spatiotemporal information via self-attention~\citep{yang2023biot} or bottleneck attention~\citep{zhou2025csbrain}.
On the other hand, \citet{ouahidi2025reve} proposes a unified framework for multiple pretraining datasets via 4D electrode positional encoding with block masking.
CodeBrain~\citep{ma2026codebrain} tokenizes both waveform and frequency for richer representation and leverages a state-space model~\citep{gu2022efficiently}.

Across these models, discrete tokenization and masked autoencoding remain central.
However, discretizing continuous EEG signals into fixed temporal units creates a structural mismatch, forcing models to reconstruct proxy representations rather than the raw waveform.
The representation operates at patch granularity, with cross-patch interactions summarized at the patch level---coarser than the time scale at which EEG dynamics actually unfold.
In contrast, we eliminate discretization entirely, modeling EEG directly in continuous time via a generative objective.

\subsection{Factorized Spatiotemporal Processing}
Decoupling temporal and spatial axes is well-established in EEG models such as EEGNet~\citep{lawhern2018eegnet} and ShallowNet~\citep{schirrmeister2017deep}, but they apply a disjoint front-end filter that immediately collapses the electrode dimension ($E \to 1$).
In contrast, our Conv(1+1)D block, inspired by R(2+1)D video convolutions~\citep{tran2018closer}, alternates temporal and electrode convolutions inside every residual block.
Together with time-only downsampling, SplitUNet preserves electrode topology and interleaves spatiotemporal mixing across the entire deep hierarchy.

\subsection{Continuous-Time Generative Models}\label{subsec:related_gen}

Continuous-time generative models learn a transport path from a simple prior to the data distribution by regressing onto a chosen interpolant.
Diffusion-based variants~\citep{ho2020denoising,karras2022elucidating,nichol2021improved} simulate a stochastic process during training, whereas flow matching~\citep{lipman2023flow} simplifies this by defining a regression target along a deterministic path, thereby removing the need for a noise schedule.
Both frameworks have been applied to neural audio waveforms~\citep{kong2021diffwave} as well as multivariate time-series imputation and forecasting~\citep{tashiro2021csdi,kollovieh2023predict,kollovieh2025flow}.
We adopt the optimal-transport conditional
flow-matching~(OT-CFM)~\citep{tong2024improving}, which pairs
each noise sample with its nearest data sample within a minibatch; the resulting velocity field is provably straighter, enabling accurate, few-step sampling without the need for distillation.

Beyond synthesis, continuous-time generative pretraining has emerged as a robust approach for representation learning.
Features from unconditional diffusion or flow models support label-efficient segmentation~\citep{baranchuk2022labelefficient,xu2023open}, dense visual correspondence~\citep{tang2023emergent}, and discriminative classification on par with contrastive
features~\citep{li2024return}.
\model extends this paradigm to multi-channel EEG waveforms, where the continuous-time generative objective serves as the foundation-model training signal.
Concurrently, EEGDM~\citep{puah2025eegdm} pretrains a structured state-space model (SSM) using DDPM diffusion~\citep{ho2020denoising} and adapts via a latent fusion Transformer.
In contrast, \model pairs optimal-transport flow matching with our hierarchical SplitUNet backbone, extracts representations directly with a simple head, and benchmarks across nine diverse downstream tasks alongside clinical fidelity evaluations.

\subsection{Time Series Foundation Model}
Time-series foundation models pretrain a single backbone on large multi-domain time-series corpora to enable zero-shot forecasting and downstream task adaptation. Approaches in this space include autoregressive next-token Transformers~\citep{ansari2024chronos,das2023decoder}, masked-encoder Transformers~\citep{woo2024unified}, and large-language-model adaptations~\citep{jin2024timellm}. While these differ in architecture, they all tokenize the input series---either into a vocabulary of quantized values or into a sequence of fixed-length patches---and the backbone operates on tokens rather than on raw continuous samples. Continuous-time generative pretraining on raw signals has not yet been adopted in this literature; diffusion- and flow-matching-based forecasting~\citep{tashiro2021csdi,kollovieh2025flow} has remained task-specific rather than foundation-model-scale.
\model differs in both respects: it operates on the raw multi-channel waveform without tokenization, and pretrains the backbone via continuous-time flow matching.


\section{Implementation Details}\label{supp:experiment_detail}

We provide additional details from Section~\ref{subsec:setup}.

\paragraph{Backbone.}
Our velocity network is a 6\,M-parameter SplitUNet adapted from the UNet of~\citet{ho2020denoising}, via the widely used \texttt{denoising-diffusion-pytorch} reimplementation\footnote{\url{https://github.com/lucidrains/denoising-diffusion-pytorch}}, with two EEG-driven modifications.
\textbf{(a) (1+1)D Factorization.}
Every $3{\times}3$ 2D convolution is replaced by a pair of 1D convolutions: a temporal convolution along the time axis followed by an electrode-mixing convolution along the electrode axis with a $(3, 1)$ kernel and $(1, 0)$ padding.
The initial input convolution uses a wider $(1, 7)$ kernel with $(0, 3)$ padding to widen the temporal receptive field.
\textbf{(b) Time-only downsampling.}
 Down- and upsampling act only along time, using $(1, 4)$ kernels at stride $(1, 2)$ with $(0, 1)$ padding, preserving the number of electrodes and their topology rather than collapsing them at the bottleneck.

The network operates on input tensors of shape $(B, 1, E, T)$ where $E$ and $T$ denote the number of electrodes and time points, with base feature width $32$, channel multipliers $(1, 2, 4, 8)$ across four encoder stages, and a mirrored decoder.
Per-stage linear-attention blocks~\citep{shen2021efficient} use $4$ heads of $32$ dimensions; a single bottleneck full-self-attention block acts where the time axis has been compressed by $8\times$.
The sinusoidal flow-time embedding ($\theta = 10{,}000$) feeds an MLP of width $128$, injected into every residual block as a per-channel scale-and-shift on the post-normalization activations~\citep{nichol2021improved}.
Self-conditioning is disabled and dropout is set to zero.

\textbf{Sampling.} Unconditional samples are drawn by integrating $\dot{\vect{x}}_t = v_\theta(\vect{x}_t, t)$ from $t{=}0$ to $t{=}1$ with a first-order (Euler) solver at $50$ uniform steps.
The straightness induced by the OT coupling makes this budget sufficient.
We did not observe further improvement in band-power $W_1$ from raising the step count.

\paragraph{Preprocessing.}
Recordings are resampled to $200$\,Hz, band-pass filtered to $0.3$--$75$\,Hz, notch-filtered at $60$\,Hz to remove signal from the US powerline, and partitioned into non-overlapping $30$\,s segments. Segments with any timepoint exceeding $100\,\mu$V in absolute amplitude are discarded; remaining segments are normalized to $[-1, +1]$ by dividing by $100\,\mu$V. We retain the $19$ channels of the international 10--20 system~\citep{klem1999ten}, excluding the ear references A1 and A2: Fp1, Fp2, F7, F3, Fz, F4, F8, T3, C3, Cz, C4, T4, T5, P3, Pz, P4, T6, O1, O2.

\paragraph{Pretraining hyperparameters.}
We optimize with Adam ($\beta_1{=}0.9$, $\beta_2{=}0.99$, no weight decay) at a constant learning rate of $8\times 10^{-5}$, with gradient clipping at $\lVert g\rVert_2 \le 1.0$.
Each step uses a batch of $4$ segments on a single NVIDIA H100 80\,GB GPU with fp16 mixed-precision training (Hugging Face Accelerate) and no gradient accumulation.
The flow time $t \sim \mathcal{U}[0,1]$ is rescaled by $1000$ before entering the time MLP.
We train for $70$ epochs and maintain an EMA of the weights with decay $0.995$ updated every $10$ optimizer steps; the EMA copy is used for all sampling and downstream feature extraction.

\paragraph{Finetuning hyperparameters.}
We optimize cross-entropy with AdamW ($\beta_1{=}0.9$, $\beta_2{=}0.999$, weight decay $0.01$) and gradient clipping at $\lVert g\rVert_2 \le 1.0$. Training uses a batch size of $32$ and runs for $40$ epochs.
For each task, we tune the learning rate ($5\cdot10^{-5}$, $10^{-4}$, or $2\cdot10^{-4}$) and the feature-aggregation strategy (mean pooling across both electrode and time axes vs.\ flattening) based on validation performance. For each run we report the test balanced accuracy of the best-performing epoch on the test split, and each cell averages the five highest-scoring seeds. The same rule is applied to every model we train ourselves for the ablation experiments (see Tables~\ref{tab:ablation_backbone}, \ref{tab:ablation_objective}, and~\ref{tab:ablation_pretrain}).

\begin{figure}[t!]
  \centering
  \includegraphics[width=\linewidth]{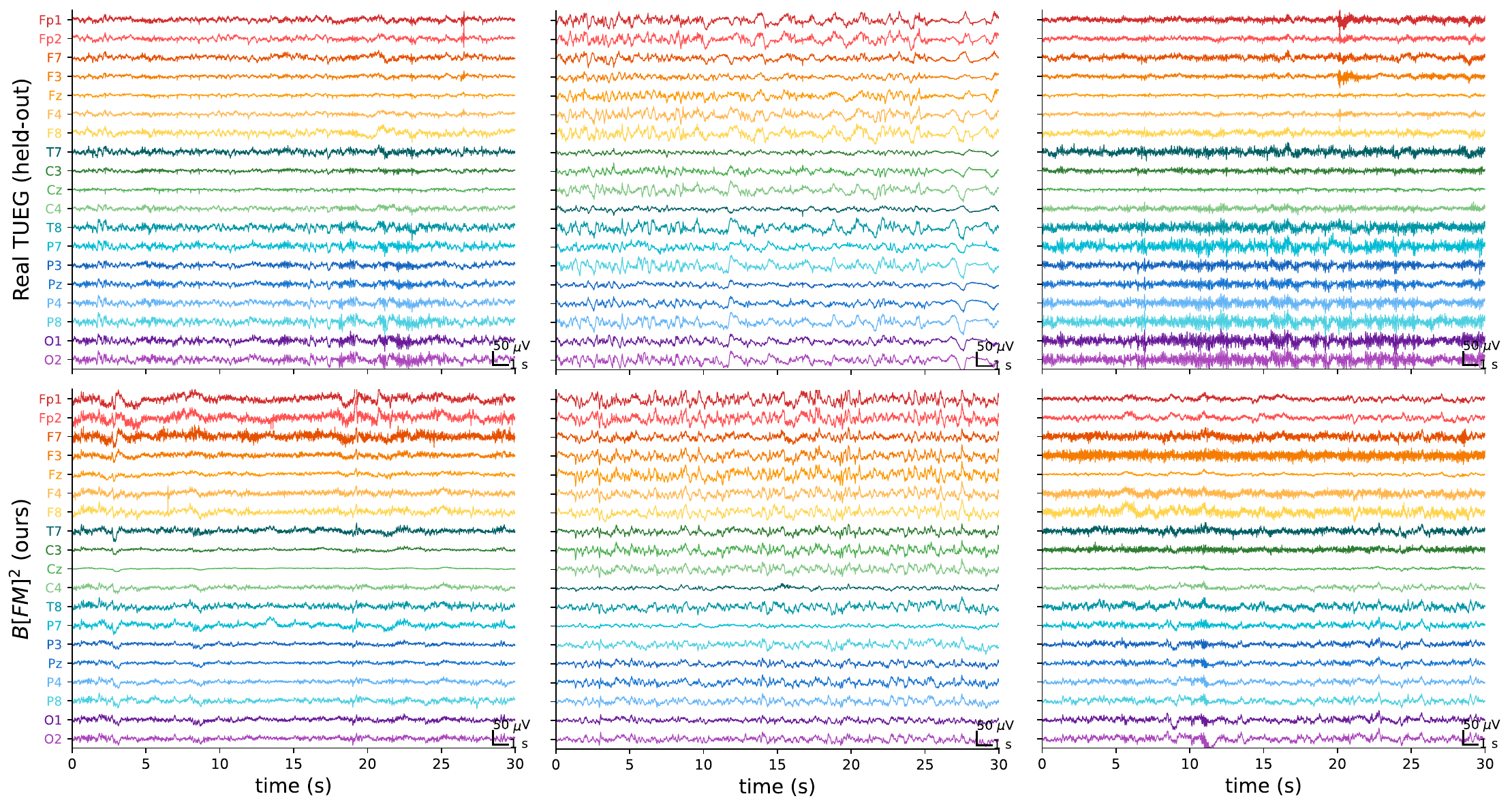}
  \caption{\textbf{Real vs.\ \model EEG segments.} Three 30\,s held-out TUEG segments (top row) and three unconditional \model samples (bottom row), shown in the $19$-channel referential 10--20 montage used for pretraining. Readers in the blinded study (see Table~\ref{tab:neurologist}; full protocol in Appendix~\ref{supp:neurologist_protocol}) received no examples in advance and viewed segments at standard clinical display settings with randomly interleaved order. The traces shown here are illustrative; subtle textural differences (e.g., apparent smoothness in synthetic samples) are illustrated in Figure~\ref{fig:gen_showcase}. Calibration bar: $1$\,s and $50\,\mu$V.}
  \label{fig:supp_samples}
\end{figure}

\section{Downstream Datasets}
\label{supp:datasets}

All recordings are resampled to $200$\,Hz before windowing. Sample counts below are rounded; precise values appear in Table~\ref{tab:downstream}.

\paragraph{TUAB.}
A binary normal-vs-abnormal benchmark drawn from the TUH EEG Corpus~\citep{obeid2016temple}, with labels assigned by clinical neurologists. We adopt the $16$-channel bipolar montage of the CBraMod~\citep{wang2025cbramod} pipelines and $10$\,s non-overlapping windows on the released subject-disjoint split: $\sim 409$k windows from $\sim 2{,}993$ patients.

\paragraph{TUEV.}
Six-class event classification from the same TUH EEG Corpus~\citep{obeid2016temple}: SPSW (spike and sharp wave), GPED, PLED, EYEM (eye movement), ARTF (artifact), BCKG (background). Same montage as TUAB; $5$\,s windows; subject-disjoint split; $\sim 112$k events. Class frequencies are strongly skewed toward BCKG.

\paragraph{BCIC-IV-2a.}
The four-class motor-imagery competition benchmark of~\citet{tangermann2012review}: $9$ subjects imagining left-hand, right-hand, both-feet, or tongue movements at $22$ channels and $250$\,Hz. After resampling and $4$\,s windowing on the released competition partition: $5{,}184$ trials.

\paragraph{ISRUC.}
Five-class AASM sleep staging from $100$ subjects in subgroup~3 of~\citet{khalighi2016isruc}. The released recordings include a chin EMG channel that CBraMod~\citep{wang2025cbramod} pipelines treat as EEG; following~\citet{ouahidi2025reve} we drop it and keep the six EEG derivations only. Subject-wise split, $30$\,s window from $\sim 89$k samples.

\paragraph{Mumtaz.}
Binary major-depressive-disorder vs.\ healthy-control classification~\citep{mumtaz2017electroencephalogram} on $19$ MDD patients and $15$ controls, $19$-channel 10--20 EEG at $256$\,Hz. The released subject-wise split with $5$\,s non-overlapping windows: $\sim 7{,}100$ samples.

\paragraph{MAT (Mental Arithmetic).}
Binary stress-vs-rest contrast~\citep{zyma2019electroencephalograms}: $36$ subjects performing timed serial subtraction against eyes-closed baseline, $20$-channel EEG at $500$\,Hz, $5$\,s windowed. The smallest task in our suite at $1{,}707$ samples and consequently the noisiest probe.

\paragraph{Siena.}
Binary seizure detection on $14$ adult patients of~\citet{detti2020eeg} ($512$\,Hz, 10--20 montage). We retain the $29$ channels available across all subjects and use $10$\,s windows ($51{,}307$ samples). Subjects PN16 and PN17 are held out as the test set; the remaining $12$ are split $8:2$ for training and validation.

\paragraph{HMC.}
Five-class AASM sleep staging from the $151$-subject PSG corpus of~\citet{alvarez2021inter}. We use the four EEG channels (F4-M1, C4-M1, O2-M1, C3-M2) at $200$\,Hz with $30$\,s epochs on a subject-wise split: $137{,}243$ samples.

\paragraph{SEED-V.}
Five-state emotion classification (happy, sad, neutral, disgust, fear) on the $16$-subject corpus of~\citet{liu2021comparing}, three sessions per subject, $62$ channels at $1000$\,Hz. We segment into $1$\,s windows on a subject-wise split: $117{,}744$ samples.

\section{Neurologist Rating Protocol}\label{supp:neurologist_protocol}

\begin{figure}[t!]
  \centering
  \begin{subfigure}[t]{0.9\textwidth}
    \centering
    \includegraphics[width=\linewidth]{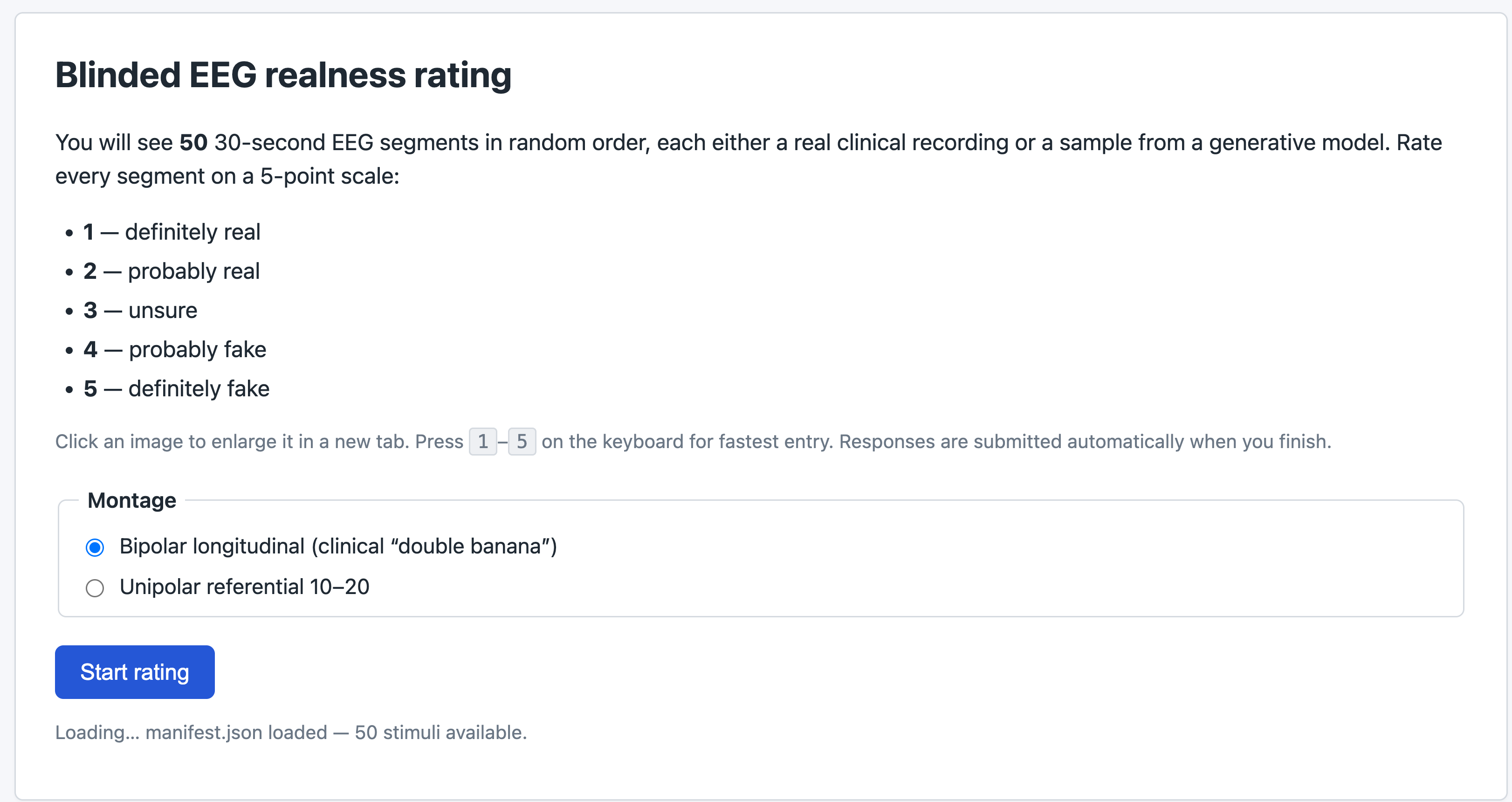}
    \caption{Welcome screen.}
    \label{fig:human_eval_begin}
  \end{subfigure}
  \\[1ex]
  \begin{subfigure}[t]{0.9\textwidth}
    \centering
    \includegraphics[width=\linewidth]{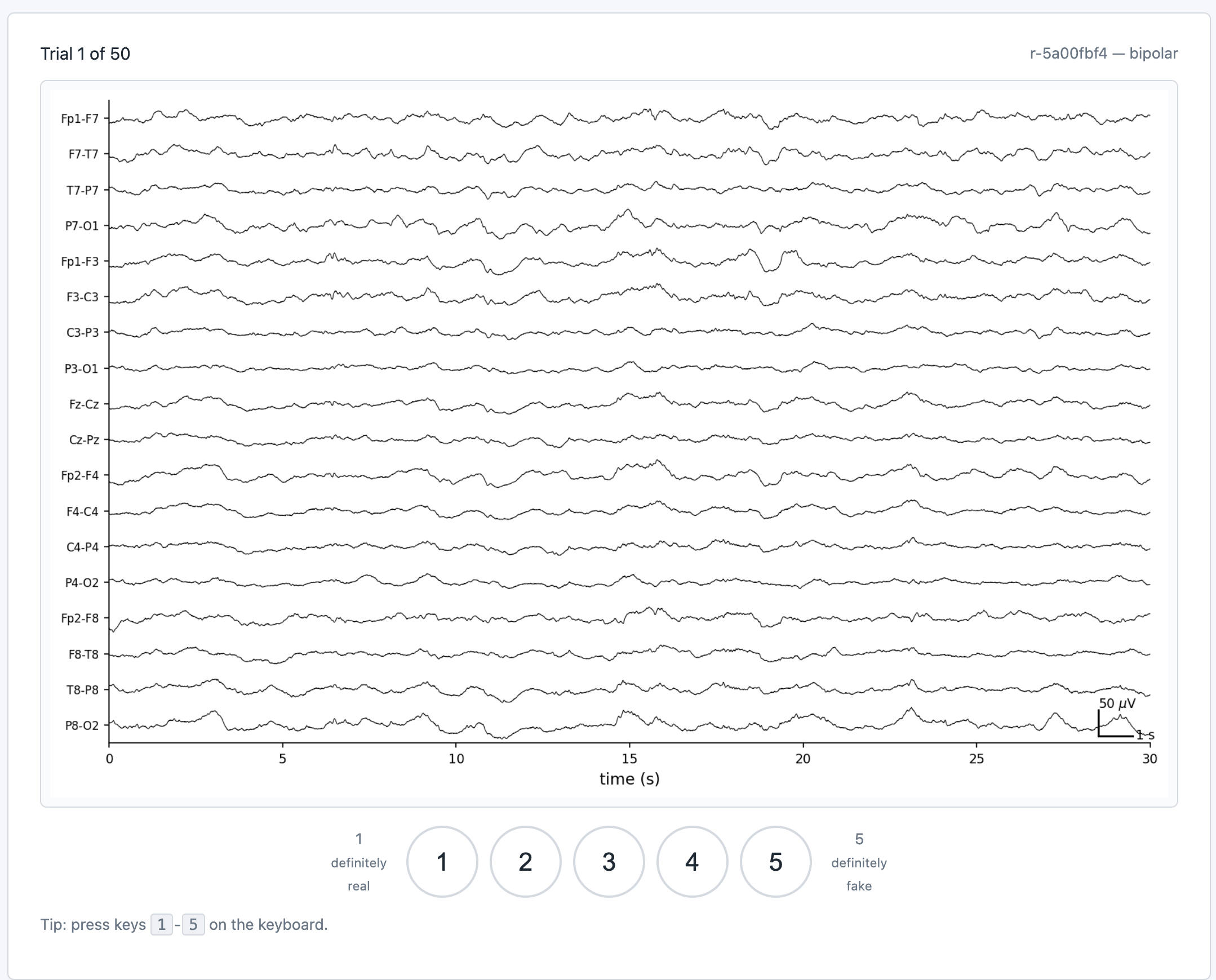}
    \caption{Per-trial rating screen.}
    \label{fig:human_eval_sample}
  \end{subfigure}
  \caption{\textbf{Blinded neurologist rating interface.}
  (a) Welcome screen describing the task, the 5-point realness Likert, and selecting the longitudinal bipolar (``double banana'') montage---the standard clinical reading layout, which removes reference-electrode artifacts and matches neurologists' habitual reading view.
  (b) Per-trial rating screen: a 30\,s segment in the bipolar montage at clinical display settings, with a 50\,$\mu$V scale bar and the Likert buttons below; trial index and an opaque stimulus identifier appear at the top so the reader cannot infer provenance.
  The interface randomly interleaves real held-out TUEG segments and unconditional \model samples and submits responses automatically; readers saw no metadata at any point.}
  \label{fig:human_eval_screens}
\end{figure}

We recruited two board-certified neurologists for the evaluation.
Stimuli were presented through a single-page web application (see Figure~\ref{fig:human_eval_screens}) at standard clinical display settings ($10$\,mm/s paper speed, $7\,\mu$V/mm sensitivity, $0.3$--$70$\,Hz bandpass, $60$\,Hz notch).
Although the original data and generated samples are in a referential montage, we convert them to a longitudinal bipolar montage to match the neurologists' routine reading convention.
Real and generated segments were matched in duration and post-preprocessing amplitude range and rendered identically; readers saw no metadata and were blind to provenance throughout.
The full Likert anchors are $1{=}$ definitely real, $2{=}$ probably real, $3{=}$ unsure, $4{=}$ probably fake, $5{=}$ definitely fake; ratings were entered by mouse or via $1$--$5$ keyboard shortcuts.

We report, per reader and pooled, the mean realness score separately for the real held-out and \model-generated subsets, and the difference between the two subset means with its Mann--Whitney $U$ test $p$ value (an unpaired test, since real and synthetic segments are independent draws).
We assess segment-level discriminability using two agreement statistics: Cohen's $\kappa$ between each reader's binarized rating (real vs.\ fake) and the true real/synthetic label, and Spearman $\rho$ between the two readers' per-segment ratings.
Samples are deemed indistinguishable from real if neither reader can reliably classify individual segments---that is, if Cohen's $\kappa$ is near zero and inter-reader Spearman $\rho$ is low---rather than from a failure to reject equality of means, which would not establish equivalence.

\section{Heuristics for Brain-State Pattern Mining for Qualitative Samples}\label{supp:heuristics}

Figure~\ref{fig:gen_showcase} shows the top-scoring generated sample for each of three brain-state patterns: a sharp-wave / spike-like transient, $\delta$-dominant slow-wave activity, and an eyes-closed posterior $\alpha$ rhythm. We mine these samples from \model's unconditional generation pool by ranking them with simple hand-crafted scores built from spectral and amplitude features. Each score is a product of (i) a \emph{shape} term---a ratio that fires when the spatial or spectral profile of a target pattern is present---and (ii) a \emph{magnitude} term $\log(1 + \cdot)$ that suppresses low-amplitude samples which happen to satisfy the shape term by chance. The pattern labels are suggestive (``$\delta$-rich''), not diagnostic.

\paragraph{Features.}
For each $E{\times}T$ generated sample (the standard 10--20 montage~\citep{klem1999ten}, $30$\,s at $200$\,Hz), we estimate the per-channel power spectral density $P(f)$ using Welch's method~\citep{welch1967use} with a 256-point Hann window and 50\% overlap, and integrate it to obtain band powers in the three canonical EEG bands relevant to our targets:
\begin{equation*}
\textstyle
b_{\delta} = \int_{0.5}^{4} P(f)\,df,\qquad
b_{\alpha} = \int_{8}^{13} P(f)\,df,\qquad
b_{\beta}  = \int_{13}^{30} P(f)\,df.
\end{equation*}
For the sharp-wave / spike-like transient pattern we use two amplitude features instead of band power: the median per-channel peak-to-peak amplitude $A_{\mathrm{pp}}$ and the count $N_{\sigma}$ of large-amplitude excursions ($\lvert z \rvert > 5$ in the per-channel z-scored signal).

\paragraph{Pattern scores.}
Let $\langle\cdot\rangle_S$ denote channel-averaging over an electrode subset $S$, where  `all' is the full 19-electrode set. The three pattern scores are:
\begin{align*}
s_{\text{spike}} &= N_{\sigma} \cdot \log\!\bigl(1+A_{\mathrm{pp}}\bigr) &&\text{(sharp-wave / spike-like transient)} \\[2pt]
s_{\text{slow}}  &= \frac{\langle b_{\delta}\rangle_{\text{all}}}{\langle b_{\delta}\rangle_{\text{all}} + \langle b_{\alpha}\rangle_{\text{all}} + \langle b_{\beta}\rangle_{\text{all}}} \cdot \log\!\bigl(1+\langle b_{\delta}\rangle_{\text{all}}\bigr) &&\text{($\delta$-dominant slow-wave activity)} \\[2pt]
s_{\text{eye}}   &= \frac{\langle b_{\alpha}\rangle_{ \{\text{O1, O2}\}}}{\langle b_{\alpha}\rangle_{\text{all}}} \cdot \log\!\bigl(1+\langle b_{\alpha}\rangle_{\text{all}}\bigr) &&\text{(posterior $\alpha$ rhythm)}.
\end{align*}
The first score fires when many large-amplitude excursions co-occur with high overall amplitude (as in interictal sharp-wave or spike-like transients); the second when low-frequency $\delta$ power dominates the spectrum (as in slow-wave sleep); and the third when $\alpha$ power is concentrated over the occipital electrodes (posterior dominant rhythm of eyes-closed wakefulness).

\section{Band-Power Distribution}\label{supp:band_power}
The five canonical bands are delta ($1$--$4$\,Hz), theta ($4$--$8$\,Hz), alpha ($8$--$13$\,Hz), beta ($13$--$30$\,Hz), and gamma ($30$--$50$\,Hz). The per-channel log-power is computed in each band, and the Wasserstein-1 distance is averaged across channels and bands to give the single value reported in Table~\ref{tab:sample-quality-arch}. Table~\ref{tab:band_power} reports the per-band breakdown.

\begin{table}[t!]
  \centering
  \small
  \setlength{\tabcolsep}{4pt}
\caption{Sample quality of \model trained with four velocity-network
backbones, measured as the Wasserstein-1 distance between
log-band-power distributions of generated and held-out real TUEG
segments ($n{=}5{,}000$ each; mean $\pm$ standard deviation across
seeds). \emph{Real-vs-Real} is a noise-floor baseline computed between
two disjoint TUEG shards. Best generator in each column (excluding
the baseline) in bold.}
\vspace{0.1cm}
  \resizebox{\linewidth}{!}{
  \begin{tabular}{lcccccc}
  \toprule
  Backbone   & $\delta$ & $\theta$ & $\alpha$ & $\beta$ & $\gamma$ & Average \\
  \midrule
  Real-vs-Real        & 0.147\,$\pm$\,0.064 & 0.147\,$\pm$\,0.029 & 0.180\,$\pm$\,0.039 & 0.165\,$\pm$\,0.076 & 0.141\,$\pm$\,0.060 & 0.156\,$\pm$\,0.028 \\
  \midrule
  UNet-1D (joint)     & 0.266\,$\pm$\,0.004 & 0.337\,$\pm$\,0.004 & 0.231\,$\pm$\,0.004 & 0.218\,$\pm$\,0.003 & \textbf{0.227\,$\pm$\,0.004} & 0.256\,$\pm$\,0.002 \\
  UNet-1D (indep.)    & 1.379\,$\pm$\,0.135 & 1.842\,$\pm$\,0.200 & 1.944\,$\pm$\,0.190 & 1.973\,$\pm$\,0.019 & 2.375\,$\pm$\,0.054 & 1.903\,$\pm$\,0.098 \\
  UNet-2D             & 0.279\,$\pm$\,0.002 & \textbf{0.158\,$\pm$\,0.003} & 0.262\,$\pm$\,0.010 & 0.230\,$\pm$\,0.009 & 0.309\,$\pm$\,0.010 & 0.248\,$\pm$\,0.001 \\
  SplitUNet         & \textbf{0.227\,$\pm$\,0.005} & 0.160\,$\pm$\,0.004 & \textbf{0.126\,$\pm$\,0.005} & \textbf{0.197\,$\pm$\,0.008} & 0.265\,$\pm$\,0.005 & \textbf{0.195\,$\pm$\,0.002} \\
  \bottomrule
  \end{tabular}
  }
  \label{tab:band_power}
  \label{tab:appbandpower}
  \end{table}

\section{Ablation Study on Feature-Extraction Timestep}
\label{subsec:supp:abl_time}

For downstream transfer, \model conditions the velocity network on $t{=}0$, the noise endpoint of the probability path.
Because a natural alternative is the clean-data endpoint ($t{=}1000$) or an intermediate flow state, we swept the conditioning timestep on TUAB across five settings spanning the interval with five seeds each (see Table~\ref{tab:tsweep}).
Downstream performance remains stable across all timesteps, with balanced accuracy spanning $0.802$--$0.819$ and AUROC $0.882$--$0.892$, a variation within typical seed-to-seed variance.
Evaluating at $t{=}0$ is therefore a practical choice rather than a sensitive hyperparameter; it achieves the strongest overall metrics while evaluating directly without simulating an extra forward step or injecting noise.

\begin{table}[t!]
  \centering
  \caption{\textbf{Sensitivity to the feature-extraction timestep on TUAB.}
  The backbone is conditioned on a fixed timestep $t$ during finetuning and inference; $t$ is on the network's internal $[0, 1000]$ scale, so $t{=}1000$ is the clean-data endpoint.
  Bold denotes the best performance.}
  \vspace{0.1cm}
  \label{tab:tsweep}
  \resizebox{0.62\linewidth}{!}{
  \begin{tabular}{lccc}
  \toprule
  $t$ & Balanced Accuracy & AUROC & AUC-PR \\
  \midrule
  $0$ (ours) & \textbf{0.819 $\pm$ 0.004} & \textbf{0.892 $\pm$ 0.008} & \textbf{0.896 $\pm$ 0.006} \\
  $250$      & 0.804 $\pm$ 0.012 & 0.882 $\pm$ 0.012 & 0.889 $\pm$ 0.012 \\
  $500$      & 0.807 $\pm$ 0.007 & 0.887 $\pm$ 0.008 & 0.892 $\pm$ 0.011 \\
  $750$      & 0.802 $\pm$ 0.005 & 0.882 $\pm$ 0.007 & 0.890 $\pm$ 0.010 \\
  $1000$     & 0.807 $\pm$ 0.007 & 0.889 $\pm$ 0.003 & \textbf{0.896 $\pm$ 0.004} \\
  \bottomrule
  \end{tabular}
  }
\end{table}

\section{Per-task detailed results}\label{supp:per_task}

For each of the nine downstream tasks, we report balanced accuracy and two task-appropriate companion metrics (Cohen's $\kappa$ and Weighted F1 for multi-class; AUC-PR and AUROC for binary). External-baseline numbers are taken directly from~\citet{ouahidi2025reve} and~\citet{zhou2025csbrain}. All \model{} entries are mean $\pm$ standard deviation over five seeds.
Dataset descriptions are in Appendix~\ref{supp:datasets}.

\subsection{Mumtaz (Mental Disorder Diagnosis)}

\begin{table}[H]
\centering
\small
\caption{Mumtaz (2-class, MDD vs.\ healthy control). External-baseline numbers from~\citet{ouahidi2025reve}.}
\label{tab:mumtaz}
\begin{tabular}{lccc}
\toprule
\textbf{Method} & \textbf{Balanced Accuracy} & \textbf{AUC-PR} & \textbf{AUROC} \\
\midrule
EEGNet           & 0.923 $\pm$ 0.010 & 0.963 $\pm$ 0.009 & 0.964 $\pm$ 0.009 \\
EEGConformer     & 0.931 $\pm$ 0.012 & 0.968 $\pm$ 0.011 & 0.970 $\pm$ 0.010 \\
SPaRCNet         & 0.932 $\pm$ 0.009 & 0.975 $\pm$ 0.006 & 0.978 $\pm$ 0.008 \\
ContraWR         & 0.919 $\pm$ 0.011 & 0.959 $\pm$ 0.010 & 0.962 $\pm$ 0.009 \\
CNN-Transformer  & 0.930 $\pm$ 0.007 & 0.976 $\pm$ 0.007 & 0.974 $\pm$ 0.006 \\
FFCL             & 0.931 $\pm$ 0.004 & 0.972 $\pm$ 0.002 & 0.975 $\pm$ 0.003 \\
ST-Transformer   & 0.913 $\pm$ 0.010 & 0.958 $\pm$ 0.009 & 0.959 $\pm$ 0.006 \\
\midrule
BIOT             & 0.936 $\pm$ 0.005 & 0.974 $\pm$ 0.003 & 0.976 $\pm$ 0.004 \\
LaBraM-Base      & 0.941 $\pm$ 0.008 & 0.980 $\pm$ 0.009 & 0.978 $\pm$ 0.006 \\
CBraMod          & 0.956 $\pm$ 0.006 & 0.992 $\pm$ 0.003 & 0.992 $\pm$ 0.003 \\
REVE        & \underline{0.964 $\pm$ 0.010} & \underline{0.996 $\pm$ 0.001} & \underline{0.996 $\pm$ 0.002} \\
CSBrain & 0.964 $\pm$ 0.015 & 0.994 $\pm$ 0.003 & 0.996 $\pm$ 0.002 \\
\midrule
\model (ours) & \textbf{\mumtazBalAccSweep} & \textbf{\mumtazAUCPRSweep} & \textbf{\mumtazAUROCSweep} \\
\bottomrule
\end{tabular}
\end{table}

\subsection{MAT (Mental Stress Detection)}

\begin{table}[H]
\centering
\small
\caption{MAT (2-class, mental arithmetic vs.\ rest). External-baseline numbers from~\citet{ouahidi2025reve}.}
\label{tab:mat}
\begin{tabular}{lccc}
\toprule
\textbf{Method} & \textbf{Balanced Accuracy} & \textbf{AUC-PR} & \textbf{AUROC} \\
\midrule
EEGNet           & 0.677 $\pm$ 0.012 & 0.576 $\pm$ 0.010 & 0.732 $\pm$ 0.011 \\
EEGConformer     & 0.680 $\pm$ 0.012 & 0.583 $\pm$ 0.013 & 0.742 $\pm$ 0.013 \\
SPaRCNet         & 0.688 $\pm$ 0.011 & 0.583 $\pm$ 0.019 & 0.742 $\pm$ 0.013 \\
ContraWR         & 0.663 $\pm$ 0.010 & 0.579 $\pm$ 0.016 & 0.733 $\pm$ 0.008 \\
CNN-Transformer  & 0.678 $\pm$ 0.027 & 0.578 $\pm$ 0.029 & 0.726 $\pm$ 0.034 \\
FFCL             & 0.680 $\pm$ 0.014 & 0.579 $\pm$ 0.027 & 0.733 $\pm$ 0.020 \\
ST-Transformer   & 0.663 $\pm$ 0.017 & 0.567 $\pm$ 0.026 & 0.713 $\pm$ 0.017 \\
\midrule
BIOT             & 0.688 $\pm$ 0.019 & 0.600 $\pm$ 0.019 & 0.754 $\pm$ 0.014 \\
LaBraM-Base      & 0.691 $\pm$ 0.013 & 0.600 $\pm$ 0.015 & 0.772 $\pm$ 0.009 \\
CBraMod          & 0.726 $\pm$ 0.013 & 0.627 $\pm$ 0.010 & 0.790 $\pm$ 0.007 \\
REVE        & \underline{0.766 $\pm$ 0.035} &   \underline{0.747 $\pm$ 0.081} & 0.845 $\pm$ 0.051 \\
CSBrain & 0.756 $\pm$ 0.011 & 0.670 $\pm$ 0.022 & \underline{0.848 $\pm$ 0.030} \\
\midrule
\model (ours) & \textbf{\matBalAccSweep} & \textbf{\matAUCPRSweep} & \textbf{\matAUROCSweep} \\
\bottomrule
\end{tabular}
\end{table}
\subsection{Siena (Seizure Detection)}

\begin{table}[H]
\centering
\small
\caption{Siena (2-class, seizure vs.\ normal). External-baseline numbers from~\citet{zhou2025csbrain}.}
\label{tab:siena}
\begin{tabular}{lccc}
\toprule
\textbf{Method} & \textbf{Balanced Accuracy} & \textbf{AUC-PR} & \textbf{AUROC} \\
\midrule
 EEGNet           & 0.749 $\pm$ 0.052 & 0.375 $\pm$ 0.087 & 0.869 $\pm$ 0.053 \\
 EEGConformer     & 0.756 $\pm$ 0.021 & 0.209 $\pm$ 0.079 & 0.816 $\pm$ 0.026 \\
 SPaRCNet         & 0.657 $\pm$ 0.038 & 0.316 $\pm$ 0.066 & 0.733 $\pm$ 0.086 \\
 ContraWR         & 0.655 $\pm$ 0.031 & 0.371 $\pm$ 0.041 & 0.782 $\pm$ 0.060 \\
 CNN-Transformer  & 0.698 $\pm$ 0.056 & 0.384 $\pm$ 0.071 & 0.872 $\pm$ 0.040 \\
 FFCL             & 0.662 $\pm$ 0.039 & 0.394 $\pm$ 0.090 & 0.815 $\pm$ 0.116 \\
 ST-Transformer   & 0.753 $\pm$ 0.038 & 0.364 $\pm$ 0.025 & 0.888 $\pm$ 0.009 \\
\midrule
 BIOT             & 0.735 $\pm$ 0.067 & 0.381 $\pm$ 0.089 & 0.903 $\pm$ 0.030 \\
 LaBraM-Base      & 0.708 $\pm$ 0.033 & 0.312 $\pm$ 0.098 & 0.881 $\pm$ 0.033 \\
 CBraMod          & 0.732 $\pm$ 0.065 & \underline{0.411 $\pm$ 0.072} & \underline{0.904 $\pm$ 0.022}\\
 REVE & 0.740 $\pm$ 0.007 & 0.381 $\pm$ 0.053 & 0.875 $\pm$ 0.030 \\ 
 CSBrain & \underline{0.766 $\pm$ 0.047} & \textbf{0.487 $\pm$ 0.034} & \textbf{0.908 $\pm$ 0.012} \\
 \midrule
 \model (ours) & \textbf{\sienaBalAcc} & \sienaAUCPR & \sienaAUROC\\
\bottomrule
\end{tabular}
\end{table}

\subsection{ISRUC (Sleep Staging)}

\begin{table}[H]
\centering
\small
\caption{ISRUC (5-class, sleep staging). External-baseline numbers from~\citet{ouahidi2025reve}. REVE numbers use the corrected six-channel montage described in Appendix~\ref{supp:datasets}.}
\label{tab:isruc}
\begin{tabular}{lccc}
\toprule
\textbf{Method} & \textbf{Balanced Accuracy} & \textbf{Cohen's $\kappa$} & \textbf{Weighted F1} \\
\midrule
EEGNet           & 0.715 $\pm$ 0.012 & 0.704 $\pm$ 0.017 & 0.751 $\pm$ 0.012 \\
EEGConformer     & 0.740 $\pm$ 0.013 & 0.714 $\pm$ 0.016 & 0.763 $\pm$ 0.015 \\
SPaRCNet         & 0.749 $\pm$ 0.007 & 0.710 $\pm$ 0.013 & 0.762 $\pm$ 0.009 \\
ContraWR         & 0.740 $\pm$ 0.013 & 0.718 $\pm$ 0.016 & 0.761 $\pm$ 0.014 \\
CNN-Transformer  & 0.736 $\pm$ 0.009 & 0.713 $\pm$ 0.012 & 0.772 $\pm$ 0.011 \\
FFCL             & 0.728 $\pm$ 0.018 & 0.702 $\pm$ 0.029 & 0.761 $\pm$ 0.020 \\
ST-Transformer   & 0.738 $\pm$ 0.021 & 0.701 $\pm$ 0.035 & 0.768 $\pm$ 0.018 \\
\midrule
BIOT             & 0.753 $\pm$ 0.012 & 0.719 $\pm$ 0.023 & 0.779 $\pm$ 0.015 \\
LaBraM-Base      & 0.763 $\pm$ 0.010 & 0.723 $\pm$ 0.018 & 0.781 $\pm$ 0.013 \\
CBraMod          & 0.786 $\pm$ 0.011 & 0.744 $\pm$ 0.015 & \underline{0.801 $\pm$ 0.010} \\
REVE        & 0.782 $\pm$ 0.008 & \underline{0.750 $\pm$ 0.016} & 0.800 $\pm$ 0.013 \\
CSBrain          & \underline{0.792 $\pm$ 0.003} & 0.741 $\pm$ 0.010 & 0.799 $\pm$ 0.009 \\
\midrule
\model (ours) & \textbf{\isrucBalAcc} & \textbf{\isrucKappa} & \textbf{\isrucFtxt} \\
\bottomrule
\end{tabular}
\end{table}
\subsection{HMC (Sleep Staging)}

\begin{table}[H]
\centering
\small
\caption{HMC (5-class, sleep staging). External-baseline numbers from~\citet{zhou2025csbrain}.}
\label{tab:hmc}
\begin{tabular}{lccc}
\toprule
 \textbf{Method} & \textbf{Balanced Accuracy} & \textbf{Cohen's Kappa} & \textbf{Weighted F1} \\
\midrule
EEGNet           & 0.653 $\pm$ 0.012 & 0.589 $\pm$ 0.020 & 0.654 $\pm$ 0.017 \\
EEGConformer     & 0.715 $\pm$ 0.009 & 0.643 $\pm$ 0.005 & 0.708 $\pm$ 0.004 \\
SPaRCNet         & 0.476 $\pm$ 0.111 & 0.315 $\pm$ 0.132 & 0.411 $\pm$ 0.131 \\
ContraWR         & 0.424 $\pm$ 0.054 & 0.234 $\pm$ 0.055 & 0.299 $\pm$ 0.029 \\
CNN-Transformer  & 0.657 $\pm$ 0.014 & 0.596 $\pm$ 0.011 & 0.690 $\pm$ 0.006 \\
FFCL             & 0.443 $\pm$ 0.070 & 0.254 $\pm$ 0.065 & 0.290 $\pm$ 0.049 \\
ST-Transformer   & 0.256 $\pm$ 0.014 & 0.050 $\pm$ 0.018 & 0.143 $\pm$ 0.012 \\
\midrule
BIOT             & 0.686 $\pm$ 0.004 & 0.629 $\pm$ 0.011 & 0.709 $\pm$ 0.015 \\
LaBraM-Base      & 0.728 $\pm$ 0.010 & 0.681 $\pm$ 0.005 & 0.745 $\pm$ 0.003 \\
CBraMod          & 0.727 $\pm$ 0.004 & 0.668 $\pm$ 0.010 & 0.740 $\pm$ 0.009 \\
REVE & \underline{0.740 $\pm$ 0.007} & \underline{0.698 $\pm$ 0.008} & \underline{0.764 $\pm$ 0.007} \\
CSBrain & 0.735 $\pm$ 0.005 & 0.682 $\pm$ 0.005 & 0.751 $\pm$ 0.004 \\
\midrule
\model (ours) & \textbf{\hmcBalAcc} & \textbf{\hmcKappa} & \textbf{\hmcFtxt} \\
\bottomrule
\end{tabular}
\end{table}

\subsection{TUEV (Event Type Classification)}

\begin{table}[H]
\centering
\small
\caption{TUEV (6-class, epileptiform event). External-baseline numbers from~\citet{ouahidi2025reve}.}
\label{tab:tuev}
\begin{tabular}{lccc}
\toprule
\textbf{Method} & \textbf{Balanced Accuracy} & \textbf{Cohen's $\kappa$} & \textbf{Weighted F1} \\
\midrule
EEGNet           & 0.388 $\pm$ 0.014 & 0.358 $\pm$ 0.015 & 0.654 $\pm$ 0.012 \\
EEGConformer     & 0.407 $\pm$ 0.016 & 0.397 $\pm$ 0.019 & 0.698 $\pm$ 0.015 \\
SPaRCNet         & 0.416 $\pm$ 0.026 & 0.423 $\pm$ 0.018 & 0.702 $\pm$ 0.010 \\
ContraWR         & 0.438 $\pm$ 0.035 & 0.391 $\pm$ 0.024 & 0.689 $\pm$ 0.014 \\
CNN-Transformer  & 0.409 $\pm$ 0.016 & 0.382 $\pm$ 0.013 & 0.685 $\pm$ 0.029 \\
FFCL             & 0.398 $\pm$ 0.010 & 0.373 $\pm$ 0.019 & 0.678 $\pm$ 0.012 \\
ST-Transformer   & 0.398 $\pm$ 0.023 & 0.377 $\pm$ 0.031 & 0.682 $\pm$ 0.019 \\
\midrule
BIOT             & 0.528 $\pm$ 0.022 & 0.527 $\pm$ 0.025 & 0.749 $\pm$ 0.008 \\
LaBraM-Base      & 0.641 $\pm$ 0.006 & 0.664 $\pm$ 0.009 & 0.831 $\pm$ 0.005 \\
CBraMod          & 0.667 $\pm$ 0.011 & 0.677 $\pm$ 0.010 & \underline{0.834 $\pm$ 0.006} \\
REVE        & 0.676 $\pm$ 0.023 & \underline{0.678 $\pm$ 0.020} & \textbf{0.845 $\pm$ 0.013} \\
CSBrain      & \underline{0.690 $\pm$ 0.006} & \textbf{0.683 $\pm$ 0.005} & 0.833 $\pm$ 0.006 \\
\midrule
\model (ours) & \textbf{\tuevBalAcc} & \tuevKappa & \tuevFtxt \\
\bottomrule
\end{tabular}
\end{table}
\subsection{TUAB (Abnormal Detection)}

\begin{table}[H]
\centering
\small
\caption{TUAB (binary normal vs.\ abnormal). External-baseline numbers from~\citet{ouahidi2025reve}.}
\label{tab:tuab}
\begin{tabular}{lccc}
\toprule
\textbf{Method} & \textbf{Balanced Accuracy} & \textbf{AUC-PR} & \textbf{AUROC} \\
\midrule
EEGNet           & 0.764 $\pm$ 0.004 & 0.830 $\pm$ 0.004 & 0.841 $\pm$ 0.003 \\
EEGConformer     & 0.776 $\pm$ 0.005 & 0.843 $\pm$ 0.005 & 0.845 $\pm$ 0.004 \\
SPaRCNet         & 0.790 $\pm$ 0.002 & 0.841 $\pm$ 0.002 & 0.868 $\pm$ 0.001 \\
ContraWR         & 0.775 $\pm$ 0.004 & 0.842 $\pm$ 0.010 & 0.846 $\pm$ 0.007 \\
CNN-Transformer  & 0.778 $\pm$ 0.002 & 0.843 $\pm$ 0.004 & 0.846 $\pm$ 0.001 \\
FFCL             & 0.785 $\pm$ 0.004 & 0.845 $\pm$ 0.006 & 0.857 $\pm$ 0.005 \\
ST-Transformer   & 0.797 $\pm$ 0.002 & 0.852 $\pm$ 0.003 & 0.871 $\pm$ 0.002 \\
\midrule
BIOT             & 0.796 $\pm$ 0.006 & 0.879 $\pm$ 0.002 & 0.881 $\pm$ 0.004 \\
LaBraM-Base      & 0.814 $\pm$ 0.002 & 0.896 $\pm$ 0.002 & 0.902 $\pm$ 0.001 \\
CBraMod          & \underline{0.829 $\pm$ 0.002} & \underline{0.926 $\pm$ 0.001} & \underline{0.923 $\pm$ 0.001} \\
REVE       & \textbf{0.832 $\pm$ 0.001} & \textbf{0.928 $\pm$ 0.001} & \textbf{0.924 $\pm$ 0.001} \\
CSBrain       & 0.817 $\pm$ 0.004 & 0.900 $\pm$ 0.007 & 0.896 $\pm$ 0.005 \\

\midrule
\model (ours) & \tuabBalAcc & \tuabAUCPR & \tuabAUROC \\
\bottomrule
\end{tabular}
\end{table}

\subsection{BCIC-IV-2a (Motor Imagery)}

\begin{table}[H]
\centering
\small
\caption{BCIC-IV-2a (4-class, motor imagery) dataset. External-baseline numbers from~\citet{ouahidi2025reve}.}
\label{tab:mi_bcic}
\begin{tabular}{lccc}
\toprule
\textbf{Method} & \textbf{Bal.\ Acc.} & \textbf{$\kappa$} & \textbf{W-F1} \\
\midrule
EEGNet          & 0.448 $\pm$ 0.009 & 0.269 $\pm$ 0.012 & 0.423 $\pm$ 0.011 \\
EEGConformer    & 0.470 $\pm$ 0.011 & 0.292 $\pm$ 0.014 & 0.453 $\pm$ 0.013 \\
SPaRCNet        & 0.464 $\pm$ 0.012 & 0.285 $\pm$ 0.015 & 0.443 $\pm$ 0.013 \\
ContraWR        & 0.468 $\pm$ 0.013 & 0.290 $\pm$ 0.016 & 0.441 $\pm$ 0.014 \\
CNN-Transformer & 0.460 $\pm$ 0.011 & 0.280 $\pm$ 0.015 & 0.446 $\pm$ 0.011 \\
FFCL            & 0.447 $\pm$ 0.014 & 0.263 $\pm$ 0.018 & 0.424 $\pm$ 0.014 \\
ST-Transformer  & 0.458 $\pm$ 0.015 & 0.273 $\pm$ 0.020 & 0.447 $\pm$ 0.014 \\
\midrule
BIOT            & 0.475 $\pm$ 0.009 & 0.300 $\pm$ 0.014 & 0.461 $\pm$ 0.013 \\
LaBraM-Base     & 0.487 $\pm$ 0.009 & 0.316 $\pm$ 0.015 & 0.476 $\pm$ 0.010 \\
CBraMod         & 0.514 $\pm$ 0.007 & 0.352 $\pm$ 0.009 & 0.498 $\pm$ 0.009 \\
REVE            & \textbf{0.640 $\pm$ 0.009} & \textbf{0.519 $\pm$ 0.013} & \textbf{0.634 $\pm$ 0.011} \\
CSBrain         & 0.566 $\pm$ 0.007 & 0.421 $\pm$ 0.009 & 0.564 $\pm$ 0.009 \\
\midrule
\model (ours) & \underline{\bcicivaBalAccSweep} & \underline{\bcicivaKappaSweep} & \underline{\bcicivaFtxtSweep} \\
\bottomrule
\end{tabular}
\end{table}

\subsection{SEED-V (Emotion Recognition)}

\begin{table}[H]
\centering
\small
\caption{SEED-V (5-class, emotion). External-baseline numbers from~\citet{zhou2025csbrain}.}
\label{tab:seedv}
\begin{tabular}{lccc}
\toprule
\textbf{Method} & \textbf{Balanced Accuracy} & \textbf{Cohen's $\kappa$} & \textbf{Weighted F1} \\
\midrule
EEGNet           & 0.296 $\pm$ 0.010 & 0.101 $\pm$ 0.014 & 0.275 $\pm$ 0.010 \\
EEGConformer     & 0.354 $\pm$ 0.011 & 0.177 $\pm$ 0.017 & 0.349 $\pm$ 0.014 \\
SPaRCNet         & 0.295 $\pm$ 0.008 & 0.112 $\pm$ 0.014 & 0.298 $\pm$ 0.008 \\
ContraWR         & 0.355 $\pm$ 0.011 & 0.191 $\pm$ 0.019 & 0.354 $\pm$ 0.012 \\
CNN-Transformer  & 0.368 $\pm$ 0.008 & 0.207 $\pm$ 0.018 & 0.364 $\pm$ 0.009 \\
FFCL             & 0.364 $\pm$ 0.009 & 0.208 $\pm$ 0.020 & 0.364 $\pm$ 0.013 \\
ST-Transformer   & 0.305 $\pm$ 0.007 & 0.108 $\pm$ 0.012 & 0.283 $\pm$ 0.011 \\
\midrule
BIOT             & 0.384 $\pm$ 0.019 & 0.226 $\pm$ 0.026 & 0.386 $\pm$ 0.020 \\
LaBraM-Base      & 0.398 $\pm$ 0.014 & 0.239 $\pm$ 0.021 & 0.397 $\pm$ 0.011 \\
CBraMod          & 0.409 $\pm$ 0.010 & 0.257 $\pm$ 0.014 & 0.410 $\pm$ 0.011 \\
REVE 		& 0.405 $\pm$ 0.002 & 0.262 $\pm$ 0.002 & 0.414 $\pm$ 0.002\\
CSBrain & \underline{0.420 $\pm$ 0.003} & \underline{0.279 $\pm$ 0.003} & \underline{0.428 $\pm$ 0.002} \\
\midrule
\textbf{\model} (ours) & \textbf{\seedvBalAcc} & \textbf{\seedvKappa} & \textbf{\seedvFtxt} \\
\bottomrule
\end{tabular}
\end{table}

\section{Compute Resources}\label{supp:compute}

Pretraining, finetuning, and sampling were run on a single NVIDIA H100 GPU with PyTorch $2.6.0$ (CUDA $12.4$) and Hugging Face Accelerate; data were served from LMDB on a shared parallel filesystem.

\section{Use of Existing Assets}\label{supp:assets}

\paragraph{Datasets.}
The Temple University Hospital EEG Corpus (TUEG) \citep{obeid2016temple} is available for research after registration at the TUH EEG portal. The downstream datasets, summarized per-dataset in Appendix~\ref{supp:datasets}, are released under the following terms: \textbf{Siena} \citep{detti2020eeg} and \textbf{MAT} \citep{zyma2019electroencephalograms} are available under the Open Data Commons Attribution License via PhysioNet; \textbf{HMC} \citep{alvarez2021inter} is provided under the Creative Commons Attribution 4.0 International license; \textbf{TUAB} and \textbf{TUEV} \citep{obeid2016temple} follow the TUH EEG terms, matching those of TUEG; \textbf{ISRUC} \citep{khalighi2016isruc} and \textbf{SEED-V} \citep{liu2021comparing} require a research-use agreement and registration; \textbf{Mumtaz} \citep{mumtaz2017electroencephalogram} is available for research use under terms listed on its repository; and \textbf{BCIC-IV-2a} \citep{tangermann2012review} is released under the Creative Commons Attribution-NoDerivatives 4.0 International license.

\paragraph{Code and libraries.}
Our implementation builds on PyTorch (BSD-3), NumPy (NumPy License), SciPy (BSD-3, used for the Hungarian assignment via \texttt{scipy.optimize.linear\_sum\_assignment}), scikit-learn (BSD), and Hugging Face Accelerate (Apache 2.0). 
The training loop and EMA scaffolding are adapted from the \texttt{denoising-diffusion-pytorch} implementation\footnote{\url{https://github.com/lucidrains/denoising-diffusion-pytorch}} (MIT). MNE-Python (BSD) is used for EEG rendering in the neurologist-rating web interface.

\paragraph{External-baseline numbers.}
External-anchor numbers in Table~\ref{tab:merged} and the per-task tables in Appendix~\ref{supp:per_task} are taken directly from \citet{ouahidi2025reve} and \citet{zhou2025csbrain}. Both source papers evaluate the publicly released checkpoints for BIOT~\citep{yang2023biot}, LaBraM~\citep{jiang2024large}, CBraMod~\citep{wang2025cbramod}, and REVE~\citep{ouahidi2025reve}; the CSBrain paper additionally evaluates its own checkpoint. All baselines are reported on the same train/val/test splits our model uses, with the single ISRUC chin-EMG correction described in Appendix~\ref{supp:datasets}.

\section{Broader Impacts}\label{supp:broader_impact}

Generating a clinically indistinguishable EEG introduces significant dual-use implications.
While high-fidelity synthetic data enables privacy-preserving sharing and robust downstream augmentation, it simultaneously lowers the barrier for fabricating medical records.

\section{LLM Usage}\label{supp:llm}
During the preparation of this manuscript, we used the Large Language Model (LLM) to assist with grammar checking, typo correction, and paraphrasing.

\end{document}